%% file: main.tex
\RequirePackage{amsmath,amssymb,float}
\PassOptionsToPackage{normalem}{ulem}
\documentclass[]{academic_template}

\input{math_commands.tex}

\usepackage{array}
\usepackage{makecell}
\usepackage{colortbl}
\usepackage{algorithm}
\usepackage{algpseudocode}
\usepackage{listings}
\usepackage{url}

\newcommand{\tablestyle}[2]{\setlength{\tabcolsep}{#1}\renewcommand{\arraystretch}{#2}\centering\footnotesize}

\setlogoheight{14mm}
\setlogospacing{5mm}
\setsjtublue
\settitlerulethickness{3pt}
\abstractboxon
\setabstractframecolor{gray}
\setabstractbgcolor{gray!10}
\setlogotolineshift{5mm}
\settoprulethickness{2.5pt}
\setbottomrulethickness{1.5pt}

\title{SpatialBrain: Spatial Intelligence Growing from Experience}
\author[1,*]{Yifan Zhou}
\author[1,*]{Ziyang Gong}
\author[2,*]{Xiaojun Ye}
\author[1]{Yifei Liu}
\author[1]{Donggang Li}
\author[1]{Xiaolong Zhou}
\author[1]{Zhihang Zhong}
\author[1,\dagger]{Xue Yang}
\affiliation[1]{Shanghai Jiao Tong University}
\affiliation[2]{Zhejiang University}
\contribution[*]{Equal contribution}
\contribution[\dagger]{Corresponding Author}
\abstract{\input{sections/abstract}}
\date{\today}
\correspondence{\email{yangxue-2019-sjtu@sjtu.edu.cn}}

\begin{document}
\maketitle

\input{sections/1_introduction}
\input{sections/2_related_work}
\input{sections/3_method}
\input{sections/4_experiments}

\input{sections/5_discussion}
\input{sections/5_conclusion}

\bibliographystyle{plainnat}
\bibliography{references}

\clearpage
\appendix
\input{sections/appendix}

\end{document}

%% file: math_commands.tex
\usepackage{amsmath,amsfonts,bm}

\def\eqref#1{equation~\ref{#1}}
\def\Eqref#1{Equation~\ref{#1}}

\def\1{\bm{1}}

\DeclareMathAlphabet{\mathsfit}{\encodingdefault}{\sfdefault}{m}{sl}
\SetMathAlphabet{\mathsfit}{bold}{\encodingdefault}{\sfdefault}{bx}{n}



%% file: sections/abstract.tex
Spatial reasoning requires more than access to geometric tools: a reasoner must decide when direct observation is sufficient and when to reconstruct, measure, or change reference frames. We present SpatialBrain, a framework that retains experience about these decisions while keeping its vision-language backbone fixed. The key distinction is between evidence about a scene and experience about how to reason with that evidence. 3DFusionMemory fuses current-scene observations into queryable spatial relations. A Spatial Skill Library provides reusable tool procedures, and Spatial Cognition associates task conditions with experience about when and how to use memory and skills. Completed trajectories are abstracted into process records and consolidated between task blocks, allowing later decisions to draw on earlier execution without transferring scene coordinates or reference answers. Across five spatial-reasoning benchmarks, SpatialBrain improves on direct answers from the same backbone on four and ties on the fifth. Component comparisons, experience-coverage curves, and traced resource decisions characterize how the framework accumulates and uses these resources.

%% file: sections/1_introduction.tex
\section{Introduction}
\label{sec:intro}

The same visual scene can pose very different reasoning problems. A visible attribute may be read directly, an object-relative direction requires a reference frame, and a distance estimate may require geometric evidence from several views. Route questions additionally require keeping track of successive movements. Spatial benchmarks expose these differences through multi-image reasoning, limited-view scene construction, dynamic spatial aptitude, and degraded observations~\cite{Yang2025mmsi,Wang2025mindcube,Jia2025omnispatial,Ray2024dynamic,Zhou2026spacedg}. A useful reasoner must therefore choose the computation as well as produce the answer. Reconstructing every scene can add unnecessary cost and noisy evidence, while answering directly can overlook information that must be integrated across views.

Spatial representations learned through training~\cite{Chen2024spatialvlm,Cheng2024spatialrgpt} and tools for perception, reconstruction, and simulation~\cite{Marsili2025visual,Chen2025geometrically,Cho2026spatialclaw,Zhu2026thinking} expand the computations available to a reasoner. Agent methods also retain reflections and reusable workflows from earlier tasks~\cite{Shinn2023reflexion,Wang2023voyager,Zhao2023expel,Wang2024agent}. Access to a resource, however, does not specify when it is useful. A procedure for computing relative directions can be reused across scenes; the coordinates on which it previously operated cannot. The challenge is to retain experience that informs resource use while grounding each answer in the current scene.

SpatialBrain separates scene evidence, reusable computation, and experience
about resource use. It operates as an agent: a vision-language reasoner that
can select and call tools. After completing tasks, it retains abstract
records of the conditions encountered, resources used, and execution
feedback. Later tasks consult these records to choose a reasoning route,
memory-access mode, or skill. We call this cross-task adaptation
\emph{self-evolution}. The model weights remain fixed; the external
experience available to the controller changes. Unlike correcting an answer
within one task, self-evolution changes the state available to subsequent
tasks.

\begin{figure*}[t]
\centering
\includegraphics[width=0.98\linewidth]{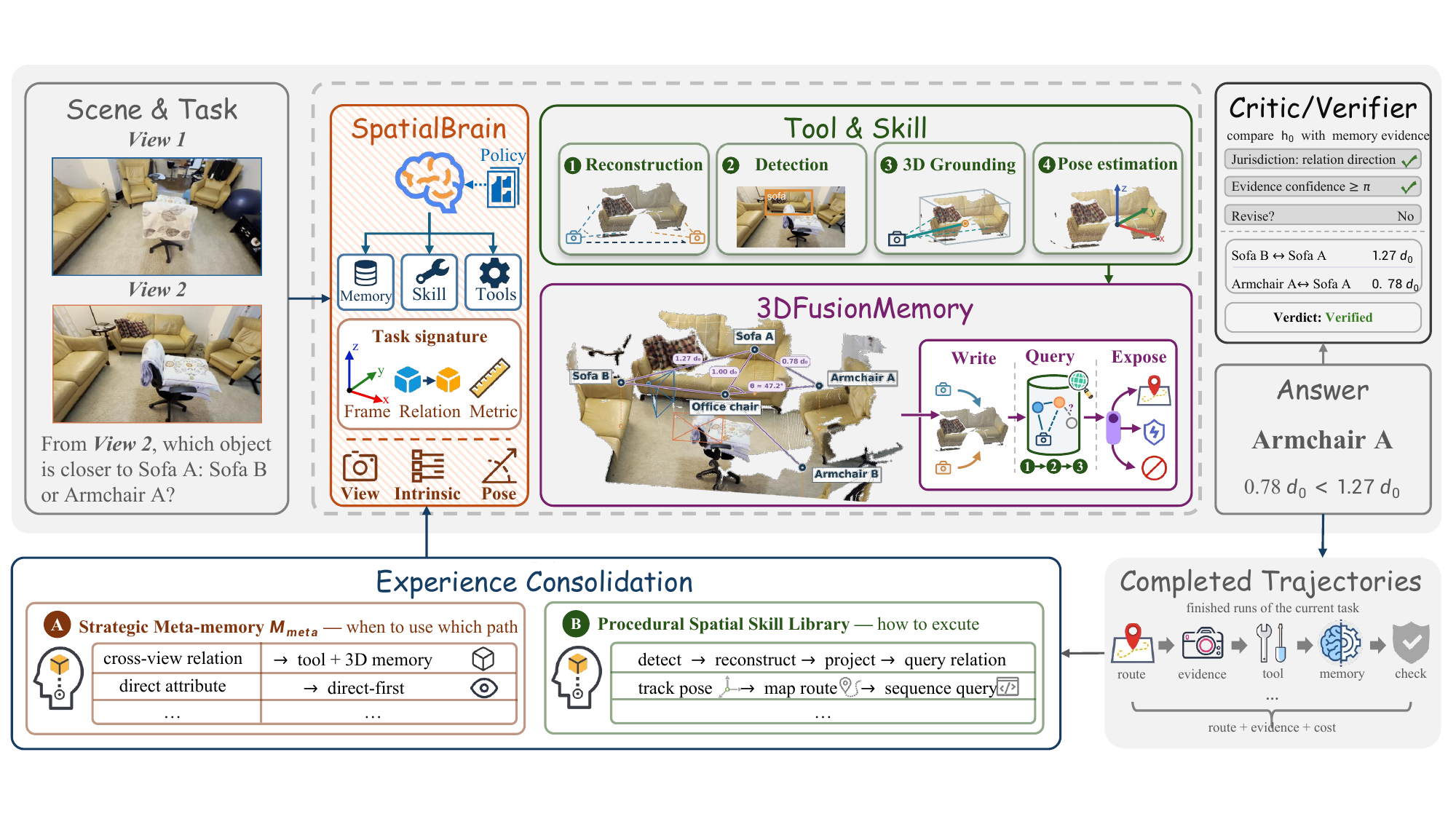}
\caption{\textbf{SpatialBrain: spatial reasoning with experience reuse.}
Spatial Cognition chooses how to use direct reasoning, spatial skills, and scene evidence. Completed trajectories update procedures and resource-use experience for later tasks; 3DFusionMemory is rebuilt for each scene. The scene is illustrative.}
\label{fig:spatialbrain-experience-loop}
\end{figure*}

\begin{samepage}
Our core contributions address three parts of this reasoning process
(Figure~\ref{fig:spatialbrain-experience-loop}):
\begin{itemize}
\item \textbf{3DFusionMemory} grounds reasoning in a shared scene state by
fusing object, pose, camera, and route observations into queryable spatial
evidence with supporting views and uncertainty.
\item \textbf{Spatial Skill Library} makes recurring spatial computations
reusable through parameterized tool procedures, separating the operation
from the scene-specific inputs on which it acts.
\item \textbf{Spatial Cognition} accumulates condition-specific experience
about when and how to use memory and skills. Past process feedback and costs
inform resource selection, connecting completed trajectories to later
reasoning decisions.
\end{itemize}
\end{samepage}
Tasks within a block share prior experience and maintain separate scene
memories. Updates occur after the block finishes, without reference answers.

Evaluation with GPT-5.6-sol shows improvements over direct answers on four
of five benchmarks and a tie on the fifth. Component comparisons and execution
traces examine resource use and how earlier
experience changes later decisions.

%% file: sections/2_related_work.tex
\section{Related Work}
\label{sec:related}

\paragraph{Spatial intelligence.}
SpatialVLM and SpatialRGPT develop spatial representations through targeted supervision~\cite{Chen2024spatialvlm,Cheng2024spatialrgpt}, while Thinking in Space studies cognitive maps for video reasoning~\cite{Yang2024thinking}. Tool-based approaches expose visual programs, geometric constraints, stateful computation, and simulation to a reasoner~\cite{Hu2024visual,Marsili2025visual,Chen2025geometrically,Cho2026spatialclaw,Zhu2026thinking}. Structured representations include S-Agent's scene and agent memories and AlloSpatial's allocentric maps~\cite{Dai2026sagent,Ruan2026allospatial}. SpatialCLI provides specialist tools, and SATURN and OmniView-Space emphasize perspective-aware reasoning~\cite{Zhou2026spatialcli,Kamali2026saturn,Li2026omniviewspace}. SpatialBrain uses execution experience to decide when and how to invoke spatial representations and computations.

\paragraph{Agent skills and memory.}
Executable skills and recurring workflows provide reusable computation, as in Voyager and Agent Workflow Memory~\cite{Wang2023voyager,Wang2024agent}. Skill-3D distills scene-aware tool workflows and failure lessons from rollouts for skill-guided inference and agentic post-training~\cite{Li2026skill3d}. SpatialBrain focuses on blockwise, reference-label-free test-time resource adaptation with fixed model weights, separating current-scene geometry from persistent resource-use experience. Episodic reflection and linked memory notes retain broader experience~\cite{Park2023generative,Xu2025amem}, whereas object-centric scene graphs and exploration snapshots retain spatial evidence~\cite{Gu2023conceptgraphs,Yang2024scene}. In SpatialBrain, the Skill Library stores \emph{how to execute} a procedure, strategic experience records \emph{when and how to use} it, and 3DFusionMemory represents \emph{what the current scene supports}.

\paragraph{Self-evolving agents.}
Self-evolving agents differ in what they update. Reflexion retains verbal feedback for subsequent trials~\cite{Shinn2023reflexion}, while ExpeL distills experience from training tasks for later inference~\cite{Zhao2023expel}. Agentic Context Engineering consolidates contextual playbooks through generation, reflection, and curation~\cite{Zhang2025ace}. Automated Design of Agentic Systems searches agent programs using an expanding archive~\cite{Hu2024adas}, while the Darwin G\"odel Machine modifies and tests agent code~\cite{Zhang2025dgm}. SpatialBrain updates external spatial strategies and procedural records from completed test-time trajectories.

%% file: sections/3_method.tex
\section{SpatialBrain}
\label{sec:method}

\subsection{Overview}
\label{subsec:overview}

\begin{figure*}[t]
\centering
\includegraphics[width=0.98\linewidth]{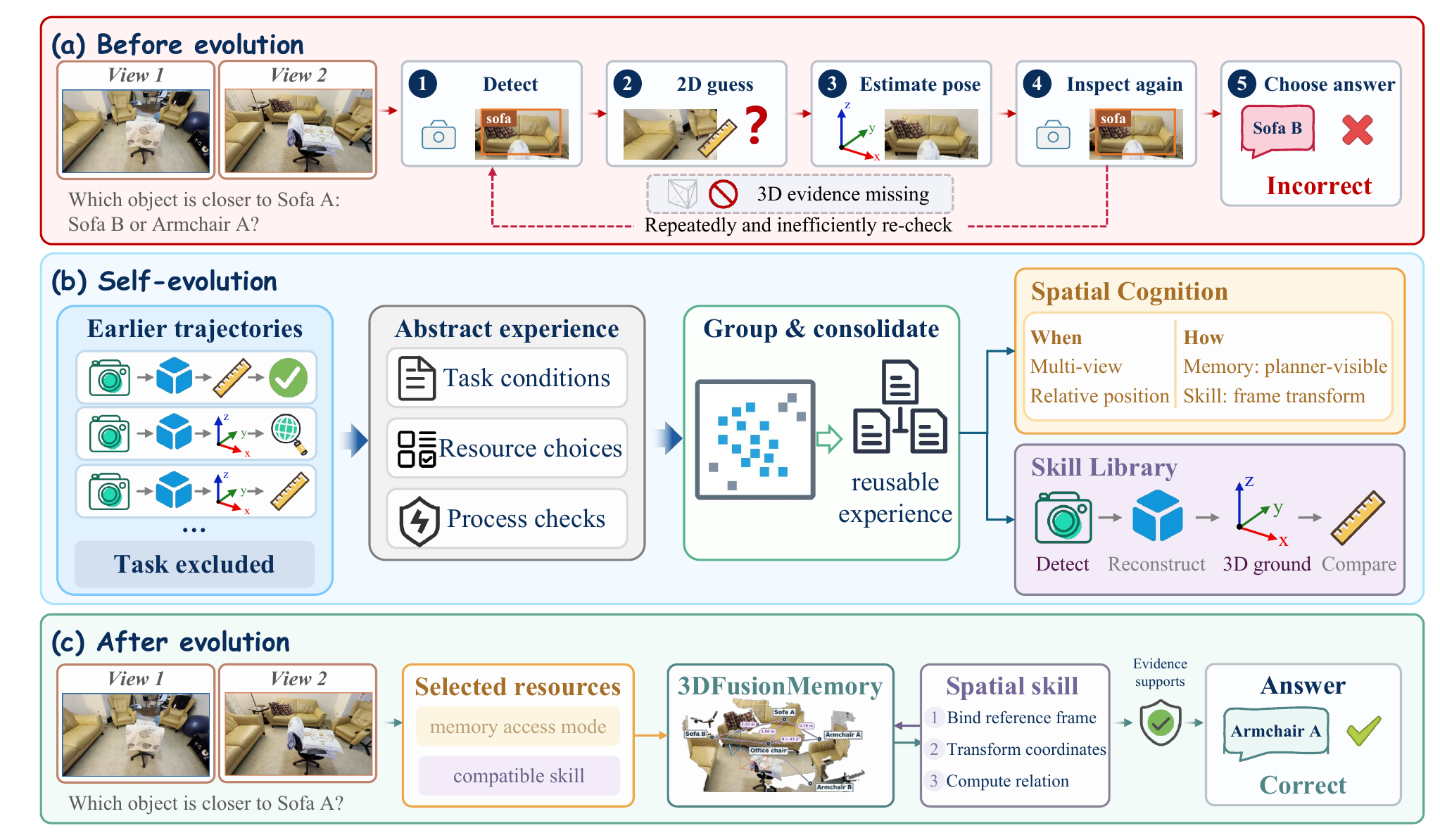}
\caption{\textbf{Self-evolution from earlier trajectories.} (a) Initial reasoning leaves the reference frame unresolved. (b) Completed earlier trajectories are abstracted and consolidated between task blocks. (c) Accumulated experience guides memory access and skill selection for the same target. The trajectories and outcomes are illustrative.}
\label{fig:architecture}
\end{figure*}

Given images $I_{1:K}$, a question $q$, and optional answer choices $O$,
SpatialBrain solves a task $x_i=(I_{1:K},q,O)$ using a fixed vision-language
backbone and perception and geometry tools. Spatial Cognition selects the
reasoning route and access to memory and skills. Tool observations populate
the scene memory, and skills carry out reusable computations on the current
inputs. An execution trajectory records these decisions, calls, returned
evidence, and process checks. Completed trajectories supply experience for
later tasks without updating model weights, as illustrated in
Figure~\ref{fig:architecture}.

This separation gives two timescales of state: \emph{scene evidence} changes
within a task, while \emph{experience about using evidence} changes between
task blocks. Let $i$ index tasks, $t$ a within-task execution step, and $b$ a
block. The state available to the reasoner is
\begin{equation}
\mathcal S_{i,t}=(\mathcal G_{i,t},\mathcal H_b),\qquad
\mathcal H_b=(\mathcal L_b,\mathcal P_b).
\label{eq:two-timescales}
\end{equation}
Here $\mathcal G_{i,t}$ is the current scene graph, $\mathcal L_b$ the skill
library, and $\mathcal P_b$ the strategic experience used by Spatial
Cognition. Observations update $\mathcal G_{i,t}$ during
execution; all tasks in a block read the same committed snapshot
$\mathcal H_b$ and keep separate scene graphs. A procedure for comparing
object-relative directions can thus persist without carrying its earlier
object coordinates into a new scene.

\subsection{3DFusionMemory}
\label{subsec:fusion-memory}

\begin{figure}[t]
\centering
\includegraphics[width=\linewidth]{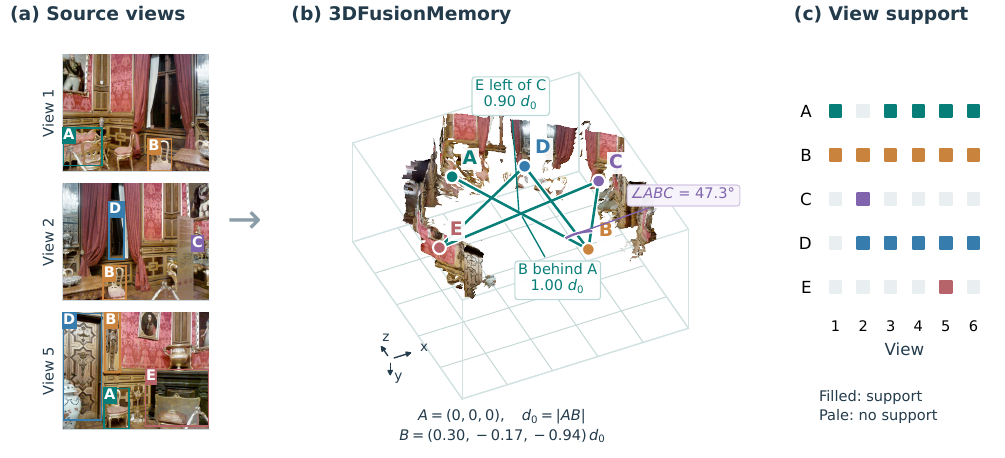}
\caption{\textbf{Multi-view observations become spatial relations.}
Three of six input views, a saved fused 3D scene, and view support for five
estimated object nodes. Colors link identities. The RGB surface uses saved
depth and camera poses; edges show unambiguous stored relations. Labels give
stored directions and 3D measurements derived from object centers.
Coordinates use A as origin and reconstruction-world axes; lengths are in
units of $d_0=|AB|$, not meters.}
\label{fig:multiview-memory}
\end{figure}

Different views expose different objects and different sides of the same
object. 3DFusionMemory fuses these observations into a shared scene frame,
where object identities connect evidence across views and spatial relations
connect objects. The memory for the current task is
\begin{equation}
\mathcal G_{i,t}=(\mathcal V_{i,t},\mathcal E_{i,t},\mathcal C_{i,t},\mathcal R_{i,t}),
\label{eq:memory-graph}
\end{equation}
where $\mathcal V_{i,t}$ contains object nodes, $\mathcal E_{i,t}$ spatial
relations, $\mathcal C_{i,t}$ camera states, and $\mathcal R_{i,t}$ ordered
route history. Nodes retain detections,
3D centers, available point sets and extents, pose, semantic axes, and
supporting views. Depth and camera poses place observations in the scene
frame; matched observations update an object's center through
confidence-weighted fusion (Appendix~\ref{app:experience-update-details}).
Repeated observations support the same node, while complementary views add
objects that a single view does not cover. Relations are then computed
between these fused object representations (Figure~\ref{fig:multiview-memory}).

Queries return distances and camera- or object-frame relations together
with confidence, supporting views, and ambiguity. Ordered route state
supports successive translations and rotations. The planner may receive
compact spatial facts, or memory may be reserved for answer checking.
Skills and reasoning steps can therefore refer to the same grounded objects
instead of reconstructing their relations independently.

\subsection{Spatial Skill Library}
\label{subsec:skill-library}

Recurring spatial operations can reuse a computation even when the objects
and images change. The library represents each skill as a procedure with an
explicit applicability condition and input--output interface:
\begin{equation}
\ell=(A_\ell,X_\ell,P_\ell,Y_\ell),\qquad \ell\in\mathcal L_b.
\label{eq:skill-interface}
\end{equation}
In \eqref{eq:skill-interface}, $A_\ell$ specifies applicability, $X_\ell$
typed input requirements, $P_\ell$ a parameterized tool procedure, and
$Y_\ell$ its returned evidence. For example, a relative-direction procedure
binds a reference frame before computing a displacement. Its parameters
refer to the current task's objects, so the same procedure can operate on
different scenes.

Seed skills initialize the library. Completed trajectories add recurring
execution patterns and records of their use. Spatial Cognition can withhold
a skill, expose it as a planning hint, or permit execution. The executable
configuration also compiles compatible primitive calls into parameterized
procedures and tests them under a limited budget; earlier hint-based
configurations retain procedural records without this compilation. A change
in an existing skill's availability is consequently distinct from the
creation of new skill code.

A skill may build or query the current scene memory, but cannot supply an
earlier answer or bypass a disabled memory interface.

\subsection{Spatial Cognition}
\label{subsec:spatial-cognition}

\paragraph{Deciding when and how to use resources.}
Spatial Cognition turns execution experience into conditional resource
decisions. Its strategic store associates input-visible task conditions
with past resource choices, process checks, failures, and costs. A task
signature $\sigma_i$ describes the reference frame, relation scope, output
form, temporal structure, metric demand, and view count. It uses neither
reference answers nor benchmark category labels. Given this signature and
the available experience, the controller selects
\begin{equation}
g_i=\pi(\sigma_i;\mathcal P_b,\mathcal L_b)
   =(r_i,m_i,\mathcal K_i,\nu_i).
\label{eq:control-policy}
\end{equation}
Here $r_i$ chooses direct-first or tool-assisted reasoning, $m_i$ controls
memory access (off, answer-checking only, or planner-visible),
$\mathcal K_i\subseteq\mathcal L_b$ gives permitted skills, and $\nu_i$
specifies verification and revision. \Eqref{eq:control-policy} is resource
selection, not answer prediction. A skill's \emph{how} describes a computation;
Cognition's \emph{how} describes how to use that computation and its evidence.
The controller combines initial rules with stored experience rather than
training an additional model.

Resource changes require supporting experience. In the executable
configuration, route and memory alternatives must have sufficient historical
support, acceptable process risk, and lower estimated token cost; otherwise
the default remains. At most one learned route or memory factor changes per
request. Skill selection requires compatible inputs and availability, while
verification and resource limits follow fixed rules.
Appendix~\ref{app:experience-update-details} specifies these conditions.
Process feedback is not a correctness reward: it describes execution
reliability and cost, not agreement with a reference answer.

\paragraph{The self-evolution loop.}
\label{subsec:evolution-method}
Tasks first execute using the fixed block snapshot. After task $i$ finishes,
its trajectory $\tau_i$ is projected to a label-free record
$d_i=\operatorname{Abs}(\tau_i)$: task signature, resource decisions, tool
sequence, process checks, failures, and available costs. This abstraction
excludes answers, literal scene coordinates, and raw task content from the
transferable state. For completed block $\mathcal B_b$, let
$\mathcal D_b=\{d_i:i\in\mathcal B_b\}$. Only then do the stores update:
\begin{equation}
\mathcal P_{b+1}=U_{\mathrm{policy}}(\mathcal P_b,\mathcal D_b),\qquad
\mathcal L_{b+1}=U_{\mathrm{skill}}(\mathcal L_b,\mathcal D_b).
\label{eq:blocked-evolution}
\end{equation}
Here $U_{\mathrm{policy}}$ consolidates condition-specific strategic records
and statistics; $U_{\mathrm{skill}}$ updates procedural records and their
availability. \Eqref{eq:blocked-evolution} is a state update, not a gradient
step. The next block can retrieve these records and make different resource
choices. No task in $\mathcal B_b$ reads these updates, future observations,
or reference answers, allowing parallel execution within each block.

%% file: sections/4_experiments.tex
\section{Experiments}
\label{sec:experiments}

\subsection{Experimental Setup}
\label{subsec:setup}

We evaluate MMSI, MindCube, OmniSpatial, SpaceDG, and SAT on 1,000, 1,050,
1,533, 1,000, and 150 items, respectively~\cite{Yang2025mmsi,Wang2025mindcube,Jia2025omnispatial,Zhou2026spacedg,Ray2024dynamic}.
We report accuracy, using mean relative accuracy (MRA) for numeric SpaceDG
questions. SpaceDG uses a fixed 1,000-item subset, and SAT uses fixed option
order. The direct-answer baseline and SpatialBrain both use GPT-5.6-sol on
the same items. Predictions are fixed before scoring; experience updates
use neither reference answers nor future tasks.

Table~\ref{tab:main} includes foundation VLMs, spatially trained models,
and agentic frameworks. It combines published results with our evaluations,
so backbones and protocols vary across methods. These comparisons position
the framework among available approaches, while the shared-backbone pair
measures its gain over direct answers. Appendix~\ref{app:full-protocols}
gives evaluation settings and comparison details.

\input{sections/spire_tables}

\subsection{Main Results}
\label{subsec:cross-benchmark}

SpatialBrain improves on direct GPT-5.6-sol answers on four benchmarks and
ties on SAT (Table~\ref{tab:main}), raising the unweighted mean from 67.5
to 72.6. MindCube accounts for the largest gain, at 20.4 percentage points;
the smaller changes on OmniSpatial and SpaceDG show that the benefit
depends on the task setting.
Category-level changes are mixed, including regressions on OmniSpatial's
egocentric questions and SpaceDG's distance and counting questions.

\subsection{Error Analysis}
\label{subsec:error-analysis}

\begin{figure}[t]
\centering
\includegraphics[width=\linewidth]{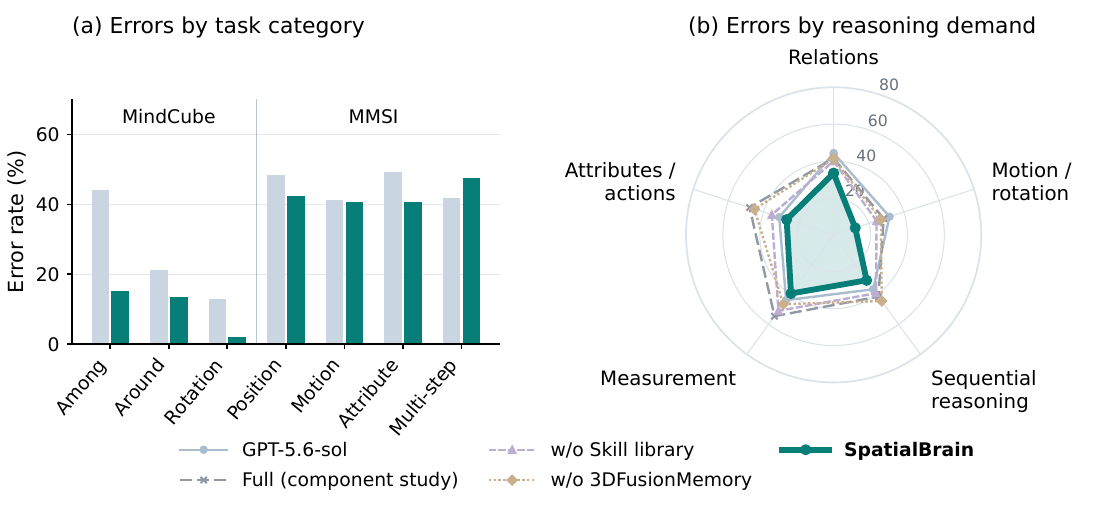}
\caption{\textbf{Errors by task type and reasoning demand.} All 1,050 MindCube
and 1,000 MMSI items. (a) Direct GPT-5.6-sol and SpatialBrain by official
category. (b) Input-derived categories for the same pair and component
configurations, all using GPT-5.6-sol. Radii show pooled error rates (0--80\%).
Appendix~\ref{app:reasoning-categories} gives the mapping, counts, and
within-study comparisons.}
\label{fig:direct-error-reduction}
\end{figure}

Figure~\ref{fig:direct-error-reduction} breaks errors down by benchmark
task type. On MindCube, errors decrease in all three categories, from 344
to 130 overall, a 62.2\% reduction. On MMSI, SpatialBrain makes 33 fewer errors
overall, but the
direction of change varies: positional and attribute errors decrease,
with no uniform reduction across all task types. The left panel groups MMSI
into its four official families. The right panel pools the two benchmarks
by input-derived reasoning demand, showing the largest reduction for
motion and rotation. These groups describe what a question requires, not
an inferred cause of failure.

The pooled radar summarizes shared demands but can hide benchmark-specific
differences: sequential-reasoning errors decrease on MindCube and increase
on MMSI. Appendix~\ref{app:direct-mmsi-errors} retains all eleven MMSI types.

\subsection{Memory-grounded Spatial Reasoning}
\label{subsec:memory-use-case}

Figure~\ref{fig:memory-case} shows two views contributing anchors to a shared
scene memory. The planner receives the atlas, and a stair-relative coordinate
check resolves the food-storage location.

\input{sections/experience_memory_case}

%% file: sections/spire_tables.tex

\begin{table*}[t]
\centering
\tablestyle{3.5pt}{1.12}
\caption{\textbf{Spatial reasoning performance across five benchmarks.}
Scores are percentages; higher is better. Our SpaceDG evaluations use a
1,000-question subset. Avg is the unweighted mean across benchmarks.
Bold marks the best displayed score; dashes indicate unavailable results.
Evaluation settings are given in Appendix~\ref{app:full-protocols}.}
\label{tab:main}
\begin{tabular}{>{\raggedright\arraybackslash}p{0.43\linewidth} ccccc c}
\toprule
Method & MMSI & MindCube & OmniSp. & SpaceDG & SAT & \textbf{Avg} \\
\midrule
\multicolumn{7}{l}{\emph{Foundation VLMs}} \\
Gemini-2.5-Pro    & 36.9 & 57.5 & 55.2 & 47.2 & 79.3 & 55.2 \\
Qwen3-VL-8B~\cite{Bai2025qwen3vl} & 27.4 & 43.4 & 51.5 & 43.1 & 74.7 & 48.0 \\
\rowcolor{gray!8}
GPT-5.5 & 51.0 & 74.3 & 67.8 & 57.5 & 84.7 & 67.1 \\
\rowcolor{gray!8}
GPT-5.6-sol & 53.8 & 67.2 & 68.3 & 56.9 & \textbf{91.3} & 67.5 \\
\midrule
\multicolumn{7}{l}{\emph{Training-based Spatial VLMs}} \\
SenseNova-SI-1.1~\cite{Cai2025sensenova} & 43.3 & 85.7 & 42.2 & 36.3 & 50.7 & 51.6 \\
SpatialLadder-3B~\cite{Li2025spatialladder} & 29.2 & 43.4 & 29.8 & 34.9 & 27.3 & 32.9 \\
RoboBrain2.5-8B-NV~\cite{Tan2026robobrain25} & 26.6 & 37.1 & 46.1 & 36.8 & 74.0 & 44.1 \\
ACE-Brain-0.5-8B~\cite{ACEBrainTeam2026acebrain05} & 35.5 & 86.3 & 41.4 & 39.8 & 82.7 & 57.1 \\
\midrule
\multicolumn{7}{l}{\emph{Agentic Frameworks}} \\
GCA~\cite{Chen2025geometrically} & 47.6 & 64.2 & 63.0 & 50.6 & 90.7 & 63.2 \\
AlloSpatial (GPT-5.5)~\cite{Ruan2026allospatial} & 43.6 & 63.2 & 63.1 & 52.9 & 89.3 & 62.4 \\
SpatialClaw (Gemma-4)~\cite{Cho2026spatialclaw} & 50.0 & 75.7 & 62.2 & 52.4 & 86.0 & 65.3 \\
Skill-3D~\cite{Li2026skill3d} & 48.8 & 69.4 & 66.6 & 51.4 & 89.3 & 65.1 \\
\midrule
\rowcolor{gray!15}
\textbf{SpatialBrain} & \textbf{57.1} & \textbf{87.6} & \textbf{69.1} & \textbf{57.9} & \textbf{91.3} & \textbf{72.6} \\
\bottomrule
\end{tabular}
\end{table*}


%% file: sections/experience_memory_case.tex
\begin{figure}[!ht]
\centering
\begingroup
\definecolor{SBCaseTeal}{HTML}{207C75}
\definecolor{SBCaseRed}{HTML}{B8493B}
\begin{tcolorbox}[enhanced,colback=white,colframe=SBCaseTeal,
colbacktitle=SBCaseTeal,coltitle=white,boxrule=.6pt,arc=1.5mm,
left=2mm,right=2mm,top=2mm,bottom=2mm,fonttitle=\bfseries\small,
title={3DFusionMemory: cross-view spatial grounding}]
\noindent
\begin{minipage}[c]{.16\linewidth}\centering
\includegraphics[width=\linewidth]{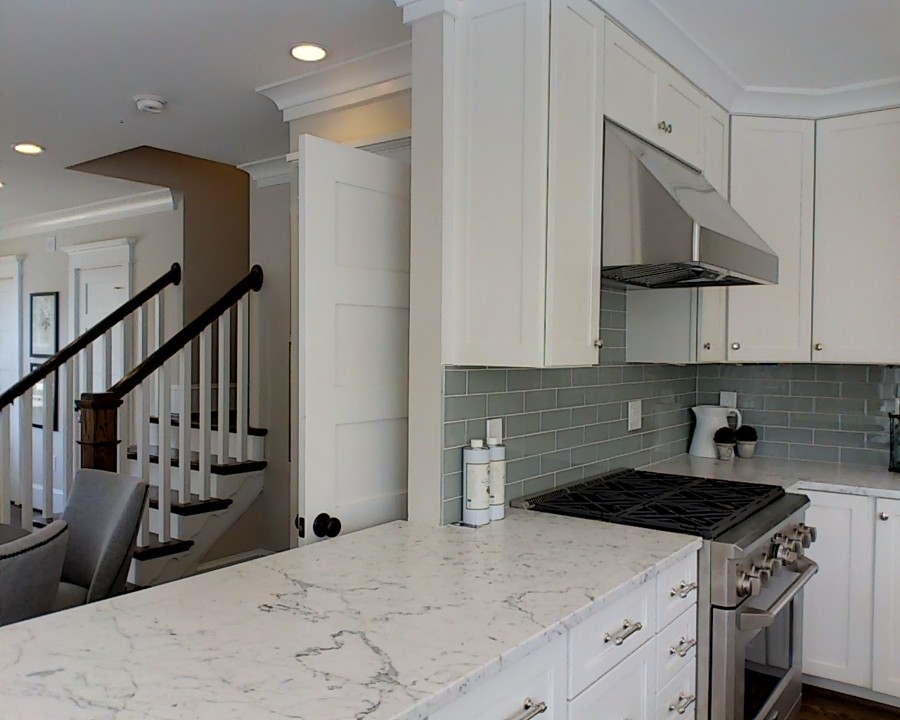}\par
{\footnotesize View 1}\par\smallskip
\includegraphics[width=\linewidth]{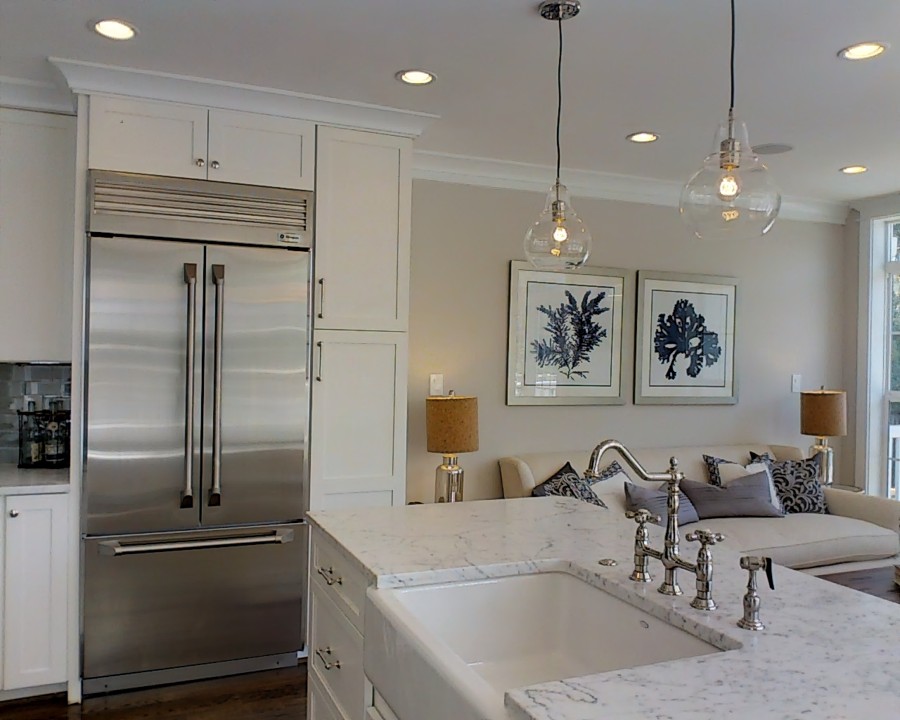}\par
{\footnotesize View 2}
\end{minipage}\hfill
\begin{minipage}[c]{.55\linewidth}\centering
{\small\bfseries\color{SBCaseTeal}3DFusionMemory}\par
\includegraphics[width=\linewidth]{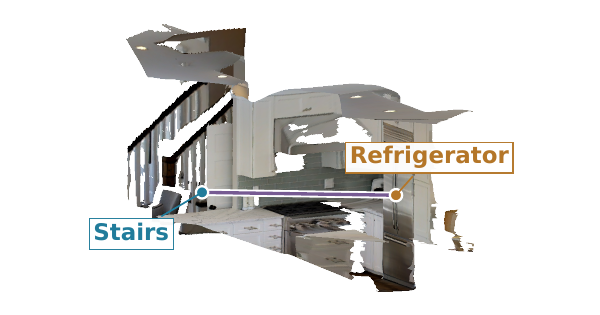}\par
{\footnotesize 2 anchors $\cdot$ 1 relation pair}
\end{minipage}\hfill
\begin{minipage}[c]{.25\linewidth}\small\raggedright
\textbf{Question}\par
Where is the food storage when going upstairs?\par\smallskip
\textcolor{SBCaseTeal}{\textbf{Coordinate check}}\par
Upstairs defines forward.\par
Right: $+1.215$\par
Forward: $+0.195$\par\smallskip
\textbf{Direct}: left\quad\textcolor{SBCaseRed}{$\times$}\par
\textcolor{SBCaseTeal}{\textbf{SpatialBrain}: right\quad$\checkmark$}
\end{minipage}
\end{tcolorbox}
\endgroup
\caption{\textbf{Memory-grounded reference-frame reasoning.} Saved 3D anchors
are shown on an RGB cutaway of the reconstructed scene. The planner sees
the memory atlas with single-view support flags; a tool verifies the
stair-relative direction. Coordinates use reconstruction units. Full
trajectory: Appendix~\ref{app:cases-memory}.}
\label{fig:memory-case}
\end{figure}

%% file: sections/5_discussion.tex
\section{Analysis and Discussion}
\label{sec:discussion}

\subsection{Component Analysis}
\label{subsec:component-results}

We compare SpatialBrain with three component-removal configurations.
\textbf{w/o Skill Library} removes seed and accumulated skills, retrieval,
hints, macros, and induction while retaining primitive tools.
\textbf{w/o 3DFusionMemory} disables scene-graph construction and access,
including queries inside skills and memory-based answer checks.
\textbf{w/o SpatialCognition} replaces task-specific analysis and adaptive
routing with a generic controller.

\begin{figure}[t]
\centering
\includegraphics[width=\linewidth]{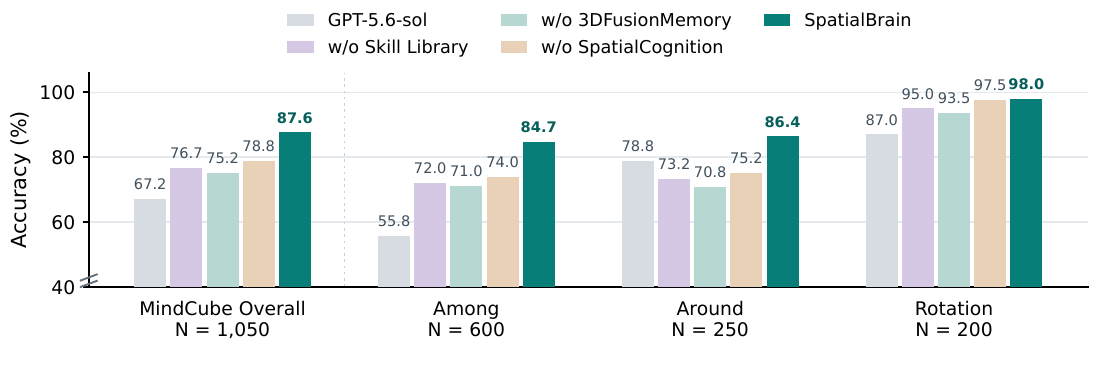}
\caption{\textbf{MindCube component comparison.} Accuracy overall and by task category on 1,050 items. The y-axis starts at 40\%. Evaluation settings are given in Appendix~\ref{app:current-component-scores}.}
\label{fig:component-results}
\end{figure}

Figure~\ref{fig:component-results} summarizes the MindCube configurations
on all 1,050 items. SpatialBrain scores 87.6, compared with 76.7 without the
Skill Library, 75.2 without 3DFusionMemory, 78.8 without SpatialCognition,
and 67.2 for direct answers. The component-removal gaps are larger on Among
and Around than on Rotation. Missing and refused answers count as wrong throughout.
Appendices~\ref{app:current-component-scores}
and~\ref{app:reasoning-categories} report the settings and answer availability.
Appendix~\ref{app:historical-mmsi-components}
reports the earlier MMSI component study (Figure~\ref{fig:mmsi-error-counts}).

\subsection{What Does 3DFusionMemory Represent?}
\label{subsec:memory-discussion}

3DFusionMemory represents spatial relations supported by complementary
observations. In Figure~\ref{fig:multiview-memory}, one view supports nodes
A and B, another adds C and D, and a third includes E.
Estimated identities connect these observations in a common scene
frame; three of the five 3D objects have support from multiple views.
The resulting memory gives later computations a shared set of objects
and relations to reference. Across five GPT-5.6-sol evaluation streams, saved 3D
objects occur in 50.6\% of MMSI tasks and 37.5\% of SpaceDG tasks, compared
with 21.0\%, 29.2\%, and 18.0\% for MindCube, OmniSpatial, and SAT
(Appendix~\ref{app:memory-coverage}). This variation reflects which inputs
produce spatial state; coverage does not measure retrieval or geometric accuracy.

\subsection{Experience and Skill Coverage}
\label{subsec:evolution-results}
\label{subsec:experience-coverage}

\label{subsec:spatialbrain-results}

\begin{figure*}[t]
\centering
\includegraphics[width=0.98\linewidth]{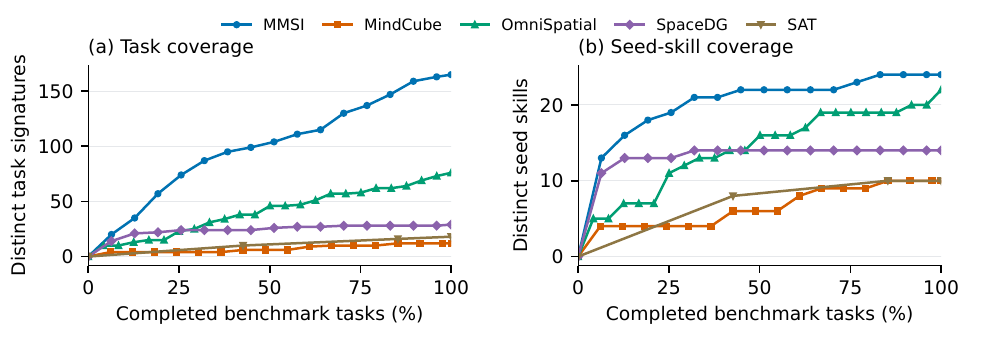}
\caption{\textbf{Test-time experience and seed-skill coverage.}
(a) Distinct task signatures with stored experience. (b) Distinct seed skills
referenced by that experience. Each curve spans the benchmark's full
evaluation scope (Section~\ref{subsec:setup}).}
\label{fig:online-accumulation}
\end{figure*}

Figure~\ref{fig:online-accumulation} tracks the breadth of stored experience
across five independent GPT-5.6-sol streams. Task coverage counts distinct
task signatures with experience records; skill coverage counts existing
seed skill IDs referenced by those records, not newly generated programs.
MMSI reaches 165 task signatures and 24 seed skills, while MindCube reaches
12 and 10. On MMSI, task coverage continues to grow after skill coverage
begins to plateau: further experience extends the recorded task conditions
without a comparable increase in distinct referenced skills.

Appendix~\ref{app:stored-experience-coverage} separately measures how often
earlier experience reaches subsequent tasks in the executable configuration.
Individual trajectories show how experience changes a route or skill
recommendation (Appendix~\ref{app:cases-routing}). An earlier GPT-5.5
hint-retrieval experiment also logged partial sequence alignment, but did
not improve aggregate accuracy or cost (Appendix~\ref{app:legacy-exposure}).
The relevant contrast for update benefits is an initial-frozen control with
the same resources and starting experience.

\paragraph{Limitations.}
\label{subsec:limitations}
Published baselines differ in protocols and serving precision, and our SAT evaluation fixes option order. Removing 3DFusionMemory also removes memory-based answer checks, so it does not isolate storage alone. The experience curves and selected trajectories establish accumulation and use, not a general improvement from self-evolution. Perception and geometry errors also remain imperfectly calibrated. Matched evaluations should measure both answer quality and the full cost of acquiring and using experience.

%% file: sections/5_conclusion.tex
\section{Conclusion}
\label{sec:conclusion}

SpatialBrain separates evidence about the current scene from reusable
spatial computation and experience about when and how to use it.
3DFusionMemory, the Spatial Skill Library, and Spatial Cognition implement
these roles around a fixed vision-language backbone. Completed trajectories
update external experience between task blocks, making self-evolution an
inspectable change in the resources and decisions available to later tasks.
The complete framework improves on direct answers on four of five
benchmarks and ties on the fifth.

%% file: sections/appendix.tex
\setcounter{figure}{0}
\renewcommand{\thefigure}{A\arabic{figure}}
\renewcommand{\theHfigure}{appendix.A.\arabic{figure}}
\setcounter{table}{0}
\renewcommand{\thetable}{A\arabic{table}}
\renewcommand{\theHtable}{appendix.A.\arabic{table}}
\setcounter{equation}{0}
\renewcommand{\theequation}{A\arabic{equation}}
\renewcommand{\theHequation}{appendix.A.\arabic{equation}}

\pdfbookmark[0]{Appendix Contents}{appendix.contents}
\section*{Appendix Contents}
\begingroup
\hypersetup{hidelinks}
\newcommand{\sbAppendixContentsLine}[2]{%
\par\noindent\hspace*{#1}%
\hyperref[#2]{\makebox[2.6em][l]{\ref*{#2}}\nameref*{#2}\dotfill\pageref*{#2}}\par}
\noindent\hfill\textbf{Page}\par\smallskip
{\bfseries\sbAppendixContentsLine{0pt}{app:full-protocols}}
\medskip
{\bfseries\sbAppendixContentsLine{0pt}{app:supplementary-quantitative}}
\smallskip
\sbAppendixContentsLine{1.5em}{app:memory-coverage}
\sbAppendixContentsLine{1.5em}{app:stored-experience-coverage}
\sbAppendixContentsLine{1.5em}{app:reasoning-categories}
\sbAppendixContentsLine{1.5em}{app:historical-mmsi-components}
\sbAppendixContentsLine{1.5em}{app:direct-mmsi-errors}
\sbAppendixContentsLine{1.5em}{app:legacy-exposure}
\medskip
{\bfseries\sbAppendixContentsLine{0pt}{app:qualitative-trajectories}}
\smallskip
\sbAppendixContentsLine{1.5em}{app:cases-memory}
\sbAppendixContentsLine{1.5em}{app:cases-skill}
\sbAppendixContentsLine{1.5em}{app:cases-routing}
\sbAppendixContentsLine{1.5em}{app:cases-tools}
\medskip
{\bfseries\sbAppendixContentsLine{0pt}{app:experience-update-details}}
\smallskip
\sbAppendixContentsLine{1.5em}{app:scene-fusion-details}
\sbAppendixContentsLine{1.5em}{app:strategy-admission-details}
\sbAppendixContentsLine{1.5em}{app:learned-fixed-state}
\endgroup

\input{sections/appendix_protocols}
\input{sections/appendix_quantitative}
\input{sections/appendix_contribution_cases}
\input{sections/appendix_experience_updates}

%% file: sections/appendix_protocols.tex
\clearpage
\section{Evaluation Settings}
\label{app:full-protocols}
Table~\ref{tab:evaluation-settings} summarizes the settings used in our evaluations.
The SpaceDG subset is sampled with seed 0 from the 9,918-item release.

\begin{table}[H]
\centering
\caption{\textbf{Evaluation settings.} The final column specifies which comparisons share each setting. Published baselines follow their original protocols.}
\label{tab:evaluation-settings}
\small
\setlength{\tabcolsep}{5pt}
\renewcommand{\arraystretch}{1.25}
\begin{tabular}{@{}>{\raggedright\arraybackslash}p{.20\linewidth}>{\raggedright\arraybackslash}p{.49\linewidth}>{\raggedright\arraybackslash}p{.24\linewidth}@{}}
\toprule
Factor & Setting & Applies to \\
\midrule
Evaluation size & MMSI 1,000; MindCube 1,050; OmniSpatial 1,533; SpaceDG 1,000; SAT 150. & Our main-table evaluations\\
Metric & Accuracy (\%); MRA for numeric SpaceDG questions. & Our scored comparisons\\
Option order & Fixed question-option order, without circular evaluation. & Our SAT evaluations\\
Backbone & GPT-5.6-sol. & Direct / SpatialBrain pair\\
Reference answers & Excluded from agent inputs and experience updates; predictions fixed before scoring. & Our agent evaluations\\
Unanswered items & Retained in the denominator and counted as wrong, including refusals. & Component comparisons\\
\bottomrule
\end{tabular}
\end{table}

{\footnotesize
Backbones, decoding settings, and tool access differ across
methods in Table~\ref{tab:main}. SenseNova-SI-1.1 denotes the InternVL3-8B variant.
AlloSpatial uses the public World2Mind tool and reasoning harness with GPT-5.5.
Qwen3-VL-8B uses a collaborator-provided score summary; raw predictions were
not available for independent verification. The World2Mind OmniSpatial result
includes three failed tool observations.}

\paragraph{MindCube component comparisons.}
\label{app:current-component-scores}
MindCube keeps all 600 Among, 250 Around, and 200 Rotation items.
The main table and error analysis use the same
920/1,050 SpatialBrain run as the Full result in
Figure~\ref{fig:component-results}. The five-stream memory and experience coverage
analyses use a separate test-time batch.
Full/no-Skill/no-Memory/no-Cognition in Figure~\ref{fig:component-results} have
0/142/135/85 blank answers, all counted wrong.
The matched Full and no-Memory comparison scores
75.9 and 75.2, respectively. Their difference is +0.67 points overall and
+2.83/-5.20/+1.50 by category; the larger gap in Figure~\ref{fig:component-results}
cannot be attributed to memory removal alone.
On 908 jointly answered Full/no-Skill items from Figure~\ref{fig:component-results},
correct counts are 799 and 805. Answer availability therefore affects the
overall gaps alongside other configuration changes.

\paragraph{MMSI component comparisons.}
\label{app:mmsi-error-counts}
Figure~\ref{fig:mmsi-error-counts} includes all four MMSI families
(Position/Motion/Attribute/multi-step: 522/150/130/198 items).
Full/no-Skill/no-Memory/no-Cognition blank counts are 68/99/116/101.
Full-minus-removal gains of 18/32/35 correct answers comprise
$-6/+12/+24$ on jointly answered pairs and $+24/+20/+11$ on pairs with a
blank. These earlier runs were not verified as complete component removals,
so their differences cannot establish isolated component effects. Task-family
breakdowns preserve the original experience stream.

%% file: sections/appendix_quantitative.tex
\clearpage
\section{Supplementary Analyses}
\label{app:supplementary-quantitative}
The Direct/SpatialBrain error comparisons use the reported main-table
predictions. The five-stream
memory and experience coverage analyses use the earlier GPT-5.6-sol
test-time batch (Appendix~\ref{app:current-component-scores}).
The multi-view illustration in Figure~\ref{fig:multiview-memory}
comes from a separate saved MMSI scene in the strict-component run. The
experience-exposure counts in Appendix~\ref{app:stored-experience-coverage}
use the later executable configuration on full MMSI and MindCube. Counts
are reported separately for each evaluation batch.

\subsection{3DFusionMemory Coverage}
\label{app:memory-coverage}

\begin{figure}[H]
\centering
\includegraphics[width=\linewidth]{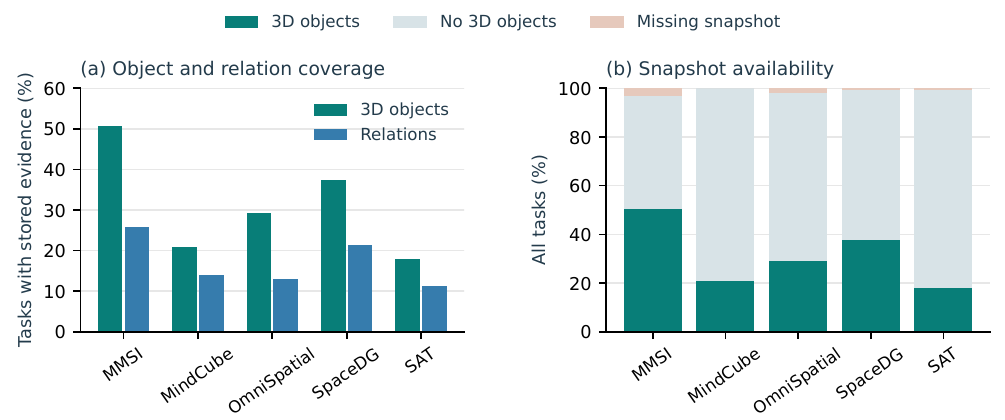}
\caption{\textbf{Stored spatial evidence across benchmarks.} (a) Tasks with
3D objects or relations. (b) Snapshot availability and 3D-object presence.
Coverage measures stored state, not retrieval or answer correctness.}
\label{fig:memory-coverage}
\end{figure}

Figure~\ref{fig:memory-coverage} counts tasks with nonempty saved 3D objects
and spatial relations. Object and relation coverage overlap, so the left
panel plots them separately. The right panel partitions each evaluation
set into snapshots with 3D objects, snapshots without 3D objects, and
missing snapshots. The last category is unknown rather than an observed
empty memory. Every percentage uses the full evaluation denominator,
including 1,000 sampled SpaceDG items.

There are 506/220/448/375/27 tasks with 3D objects and
257/146/201/213/17 with relations on MMSI/MindCube/OmniSpatial/SpaceDG/SAT.
Missing snapshots number 33/0/30/7/1. Multiple camera states occur in
710/496/0/574/0 tasks, respectively. MMSI and SpaceDG show higher object
and relation coverage than the other streams, but coverage does not follow
a universal 3D-versus-2D split. The example in
Figure~\ref{fig:multiview-memory} illustrates complementary view support:
the displayed views support two, three, and four of its five 3D objects,
and their union covers all five. This describes observed support, not
object recall against ground truth.

\subsection{Experience Coverage and Subsequent Exposure}
\label{app:stored-experience-coverage}

Beyond the stored coverage in Figure~\ref{fig:online-accumulation}, we measure
exposure to earlier experience in the executable configuration on complete
MMSI and MindCube streams. MMSI forms 18 experience records, six of which
become available; MindCube forms five, with three available. These are
condition-specific resource-use records, not new algorithms. Nonempty
experience from an earlier committed state reaches 231 of 1,000 MMSI
tasks and 212 of 1,050 MindCube tasks. Matching the injected record IDs
to each target's input state verifies that their supporting events precede
the target block. These exposure counts establish delivery to the reasoner,
not successful execution or an improvement in answers.

\subsection{Input-Derived Reasoning Categories}
\label{app:reasoning-categories}

The radar in Figure~\ref{fig:direct-error-reduction} uses input-derived
task signatures recorded by the label-blind router, with the same label
assigned to every method's response for each item. We group \texttt{relation} as
Relations; \texttt{motion} and \texttt{rotation} as Motion/rotation;
\texttt{route\_sequence} as Sequential reasoning; \texttt{visual\_measurement},
\texttt{count}, \texttt{diagram\_geometry}, and \texttt{depth} as Measurement;
and \texttt{attribute}, \texttt{appearance}, \texttt{affordance}, and
\texttt{spatial\_physics} as Other attributes/actions. This mapping was fixed
before aggregating outcomes and retains every item. Direct responses often
lack the traces needed to diagnose the cause of a wrong answer.
Table~\ref{tab:reasoning-categories} reports the category counts and pooled
error rates used in the radar.

\begin{table}[H]
\centering
\small
\caption{\textbf{Errors within input-derived categories.} Top:
Direct/SpatialBrain wrong-answer counts, with category size in parentheses.
Bottom: pooled error rates (\%) for all radar curves. Each method uses the same
2,050 items and fixed category assignment.}
\label{tab:reasoning-categories}
\begin{tabular}{lrrr}
\toprule
Category & MindCube & MMSI & Pooled \\
\midrule
Relations & 178/98 (459) & 279/247 (574) & 457/345 (1,033) \\
Motion/rotation & 117/7 (405) & 64/63 (163) & 181/70 (568) \\
Sequential reasoning & 46/25 (153) & 51/56 (113) & 97/81 (266) \\
Measurement & --- & 49/44 (112) & 49/44 (112) \\
Other attributes/actions & 3/0 (33) & 19/19 (38) & 22/19 (71) \\
\bottomrule
\end{tabular}
\medskip

\setlength{\tabcolsep}{5pt}
\begin{tabular}{lrrrrr}
\toprule
Method & Relations & Motion & Sequential & Measure. & Other \\
\midrule
GPT-5.6-sol & 44.2 & 31.9 & 36.5 & 43.8 & 31.0 \\
Full (component study) & 41.5 & 28.5 & 41.4 & 54.5 & 47.9 \\
w/o Skill library & 40.0 & 24.5 & 39.1 & 50.9 & 35.2 \\
w/o 3DFusionMemory & 41.2 & 26.9 & 44.4 & 46.4 & 45.1 \\
SpatialBrain & 33.4 & 12.3 & 30.5 & 39.3 & 26.8 \\
\bottomrule
\end{tabular}
\end{table}

The pooled error rate is lower for all five groups, while the contribution
of each benchmark differs. Motion/rotation improves mainly on MindCube;
sequential reasoning regresses on MMSI, and its Other group is unchanged.
Measurement has no MindCube items under this signature taxonomy.

The component curves come from the later executable implementation and
include its own Full reference. On MindCube/MMSI, this Full makes 253/543
errors, compared with 238/500 without the Skill library and 260/521 without
3DFusionMemory. The gap to main-table
SpatialBrain also includes implementation changes and cannot isolate a
component's contribution. MindCube's component runs passed their runtime
checks; the MMSI curves are independently scored complete predictions whose
runtime audits remain unverified. Blank answers count as errors throughout.

\clearpage
\subsection{MMSI Component Comparison}
\label{app:historical-mmsi-components}
\begin{figure}[H]
\centering
\includegraphics[width=0.90\linewidth]{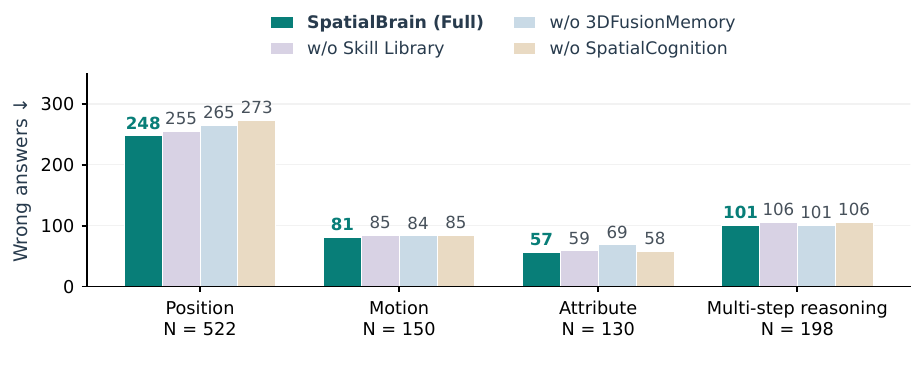}
\caption{\textbf{MMSI component errors by task family.}
Earlier component configurations on all 1,000 items. Bars count wrong
answers, including unanswered questions.
The four families partition the evaluation set; configuration and
answer-availability differences are detailed in
Appendix~\ref{app:mmsi-error-counts}.}
\label{fig:mmsi-error-counts}
\end{figure}

This earlier GPT-5.6-sol study compares Full with three removal configurations
on the same 1,000 MMSI items (Figure~\ref{fig:mmsi-error-counts}). It is
separate from the main-table comparison.
Full makes 487 errors, versus 505/519/522 for Skill/Memory/Cognition removal.
Errors differ most on positional questions; multi-step errors are unchanged
by memory removal.
The configurations were not verified as complete removals, and unanswered
items account for part of the gaps (Appendix~\ref{app:mmsi-error-counts}).

\clearpage
\subsection{MMSI Errors by Task Type}
\label{app:direct-mmsi-errors}
\begin{figure}[H]
\centering
\includegraphics[width=\linewidth]{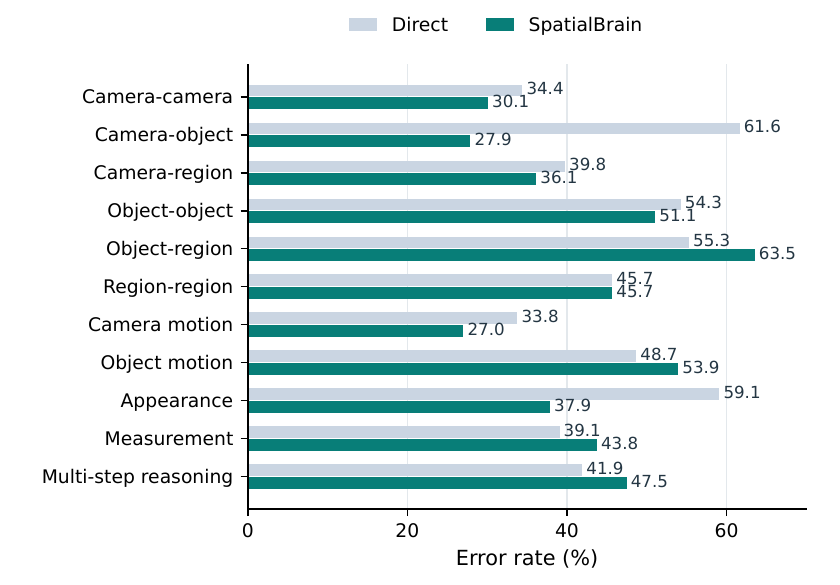}
\caption{\textbf{MMSI errors by task type.} Direct GPT-5.6-sol answers and
SpatialBrain on all 1,000 items. Each rate divides wrong answers by the
number of items in that type. Lower is better.}
\label{fig:mmsi-direct-errors}
\end{figure}

Figure~\ref{fig:mmsi-direct-errors} compares Direct and SpatialBrain on all
1,000 MMSI items and eleven task types. Direct makes 462 errors and SpatialBrain makes 429, with 199
corrections and 166 reversals. Camera--object errors decrease from 53 to 24
and appearance errors from 39 to 25, whereas multi-step errors increase
from 83 to 94. Region--region errors are unchanged.

\subsection{Earlier Retrieval and Sequence Alignment}
\label{app:legacy-exposure}
In an earlier GPT-5.5 experiment, 500 MMSI source tasks produce 81
procedural records across 36 task--path families and 16 capability
skeletons. Among 500 disjoint target tasks, 332 are eligible for retrieval:
122 retrieve a record (36.7\%) and 59 execute a partially matching tool
sequence (17.8\%). Partial alignment requires ordered overlap covering at
least half of a retrieved sequence; it occurs in 48.4\% of tasks with
retrieved records. This analysis does not count newly generated skills or
measure answer improvements. The experiment did not improve aggregate
accuracy or cost. It is separate from the full executable-cohort exposure
counts in Appendix~\ref{app:stored-experience-coverage}.

%% file: sections/appendix_contribution_cases.tex
\clearpage
\section{Qualitative Analysis and Visualizations}
\label{app:qualitative-trajectories}
\label{app:categorized-corrections}
We group the examples by their role in spatial reasoning: \textbf{3DFusionMemory}
for grounding, the \textbf{Skill Library} for reusable procedures, and
\textbf{SpatialCognition} for experience-guided resource choices. Teal,
purple, and amber boxes identify these three groups; blue-gray boxes show
additional tool-based trajectories.

\begin{center}\small
\begin{tabular}{@{}lp{.52\linewidth}@{}}
\toprule
Component / box color & Examples (MMSI IDs) \\
\midrule
\textcolor[HTML]{207C75}{3DFusionMemory / teal} & Atlas exposure and spatial verification: 9, 206; multi-view representations: 297, 861. \\
\textcolor[HTML]{65428A}{Skill Library / purple} & Successful skill returns and answer correction: 32, 42. \\
\textcolor[HTML]{946316}{SpatialCognition / amber} & Experience-guided routing: 748; adaptive skill selection: 348. \\
\textcolor[HTML]{486A83}{Supporting traces / blue-gray} & Coordinate and temporal reasoning: 18, 82; shorter paths and fallback: 289, 880; boundary cases: 715, 416. \\
\bottomrule
\end{tabular}\end{center}
\paragraph{Case selection.}
The ten detailed trajectories and two compact boundary cases come from the
1,000-item MMSI evaluations with GPT-5.6-sol. Seven show an incorrect Direct answer and a correct SpatialBrain answer. We
selected these examples after scoring to explain distinct execution paths;
they are not representative of the full set, which has 199 corrections and
166 reverse outcomes. Overall scores are 53.8\% for Direct, 57.4\% for the
fixed-policy control, and 57.1\% for SpatialBrain.

The boxes show observed operations rather than isolated component effects;
executing an existing skill is distinct from learning a new one.
The two memory visualizations illustrate saved spatial representations,
independently of answer correctness.
For the SpatialCognition cases, we checked that the cited experience precedes
the target task and that the later resource choice follows that experience.

\paragraph{Compute comparisons.}
Unless stated otherwise, compute comparisons use the fixed-policy control.
Of 485 both-correct
fixed-policy/SpatialBrain pairs, 480 have complete token records, 221 use fewer
SpatialBrain tokens, and 71 also use fewer non-final top-level calls and less
observed time. Of 372 both-correct Direct/SpatialBrain pairs, 371 have complete
tokens and only MMSI 748 uses fewer SpatialBrain tokens; both paths use zero
tools. Call counts include failed attempts and exclude final-answer generation;
primitive calls inside a skill are counted separately from top-level calls.
Elapsed times are observational: concurrency and caches were not controlled.
The evaluations used the same task IDs, model, and protocol, but exact image-byte
identity could not be checked because per-run hashes were not recorded.

\begingroup
\newcounter{sbCase}
\definecolor{SBGreen}{HTML}{147F77}
\definecolor{SBRed}{HTML}{B8493B}
\definecolor{SBCaseAccent}{HTML}{207C75}
\newcommand{\sbCaseAnswer}[4]{%
\par\noindent\begin{tabular}{@{}p{.17\linewidth}@{}p{.63\linewidth}@{}p{.20\linewidth}@{}}
\textcolor{SBCaseAccent}{\textbf{#1}} & #2 & \raggedleft\textcolor{#3}{#4}\tabularnewline
\end{tabular}\par}
\newtcolorbox{sbCompactCase}[1]{enhanced,breakable,
colback=SBCaseAccent!3!white,colframe=SBCaseAccent,colbacktitle=SBCaseAccent,
coltitle=white,boxrule=.7pt,arc=2mm,left=3mm,right=3mm,top=2mm,bottom=2mm,
fonttitle=\bfseries,fontupper=\small,title={#1}}
\newcommand{\sbCaseQuestion}{\textbf{\emph{\textcolor{SBCaseAccent}{Question}}}\par}
\newcommand{\sbCaseStep}{\par\smallskip\textcolor{SBCaseAccent}{\textbf{SpatialBrain}}\par}

\clearpage
\subsection{3DFusionMemory}
\label{app:cases-memory}

\begingroup
\definecolor{SBMechanism}{HTML}{207C75}
\refstepcounter{sbCase}
\begin{tcolorbox}[enhanced,breakable,colback=SBMechanism!3!white,colframe=SBMechanism,
colbacktitle=SBMechanism,coltitle=white,boxrule=.7pt,arc=2mm,left=3mm,right=3mm,top=2mm,bottom=3mm,fonttitle=\bfseries,
title={3DFusionMemory: Cross-view Spatial Grounding\hfill\normalfont\small MMSI 9}]
\label{case:mechanism-9}
{\footnotesize\textcolor{SBMechanism}{\textbf{3DFusionMemory}} $\cdot$ Answer Correction}\par\smallskip
{\bfseries\itshape\color{SBMechanism}Question}\par
When you are going upstairs in the stairwell area, where is the food storage area located relative to you?\par\smallskip
{\small A. In front\quad B. Behind\quad C. On the left\quad D. On the right}\par\medskip
\begin{minipage}[t]{.485\linewidth}\centering
\includegraphics[width=\linewidth,height=4.2cm,keepaspectratio]{figures/content/categorized_cases/case-9/9_0.jpg}\par
{\footnotesize Image 1}\end{minipage}\hfill
\begin{minipage}[t]{.485\linewidth}\centering
\includegraphics[width=\linewidth,height=4.2cm,keepaspectratio]{figures/content/categorized_cases/case-9/9_1.jpg}\par
{\footnotesize Image 2}\end{minipage}
\par\medskip
\textbf{Direct}\quad C. On the left\hfill\textcolor[HTML]{B8493B}{$\times$ Incorrect}\par
\smallskip\textcolor{SBMechanism!30}{\rule{\linewidth}{.35pt}}\par
\textbf{\textcolor{SBMechanism}{SpatialBrain}}\par\smallskip
\begin{minipage}[t]{.46\linewidth}\vspace{0pt}\centering
\includegraphics[width=\linewidth,height=4.2cm,keepaspectratio]{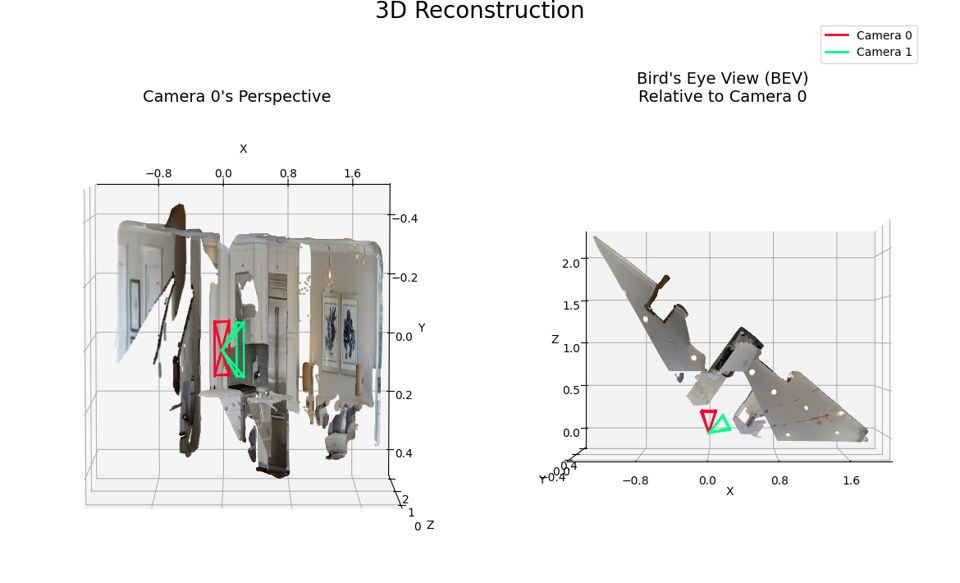}\par
{\footnotesize Reconstruction and camera poses}\par\smallskip
\begin{tcolorbox}[colback=white,colframe=SBMechanism!35,boxrule=.4pt,arc=1mm,left=2mm,right=2mm,top=1mm,bottom=1mm]
{\footnotesize \textbf{Memory available to the planner}\par 2 anchors; 1 relation pair.\par Atlas includes relation uncertainty.}
\end{tcolorbox}
\end{minipage}\hfill\begin{minipage}[t]{.50\linewidth}\vspace{0pt}
\begin{enumerate}\setlength{\itemsep}{4pt}\setlength{\parskip}{0pt}\setlength{\topsep}{0pt}
\item Reconstruct both views; detect and project the stairs and refrigerator into the shared frame.
\item The saved 3DFusionMemory atlas contains two 3D anchors and one relation and is exposed to the planner.
\item The coordinate check uses the upstairs direction as forward: right = +1.215, forward = +0.195. Return D: on the right.
\end{enumerate}\end{minipage}\par\medskip
\textbf{Answer}\quad D. On the right\hfill\textcolor[HTML]{147F77}{$\checkmark$ Correct}\par
\smallskip{\footnotesize\itshape A tool verifies the direction after atlas exposure; there is no standalone MemoryQuery call.}\par
\end{tcolorbox}
\endgroup

\clearpage

\begingroup
\definecolor{SBMechanism}{HTML}{207C75}
\refstepcounter{sbCase}
\begin{tcolorbox}[enhanced,breakable,colback=SBMechanism!3!white,colframe=SBMechanism,
colbacktitle=SBMechanism,coltitle=white,boxrule=.7pt,arc=2mm,left=3mm,right=3mm,top=2mm,bottom=3mm,fonttitle=\bfseries,
title={3DFusionMemory: Landmark-based Grounding\hfill\normalfont\small MMSI 206}]
\label{case:mechanism-206}
{\footnotesize\textcolor{SBMechanism}{\textbf{3DFusionMemory}} $\cdot$ Answer Correction}\par\smallskip
{\bfseries\itshape\color{SBMechanism}Question}\par
The stove is on the east side of the kitchen. In which direction is the mural in the kitchen?\par\smallskip
{\small A. Northeast\quad B. Southeast\quad C. Southwest\quad D. Northwest}\par\medskip
\begin{minipage}[t]{.485\linewidth}\centering
\includegraphics[width=\linewidth,height=4.2cm,keepaspectratio]{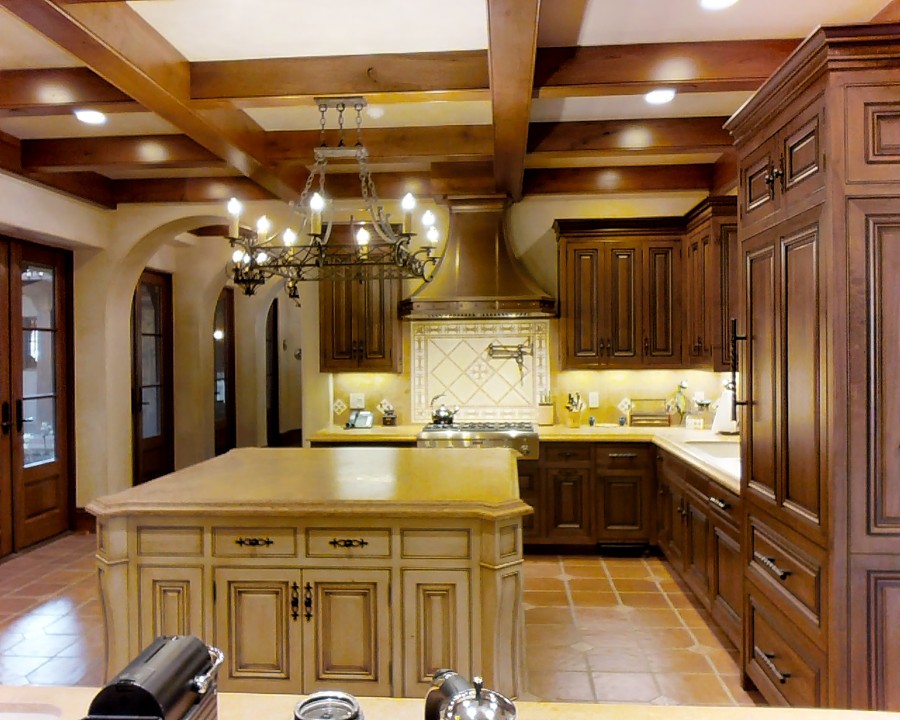}\par
{\footnotesize Image 1}\end{minipage}\hfill
\begin{minipage}[t]{.485\linewidth}\centering
\includegraphics[width=\linewidth,height=4.2cm,keepaspectratio]{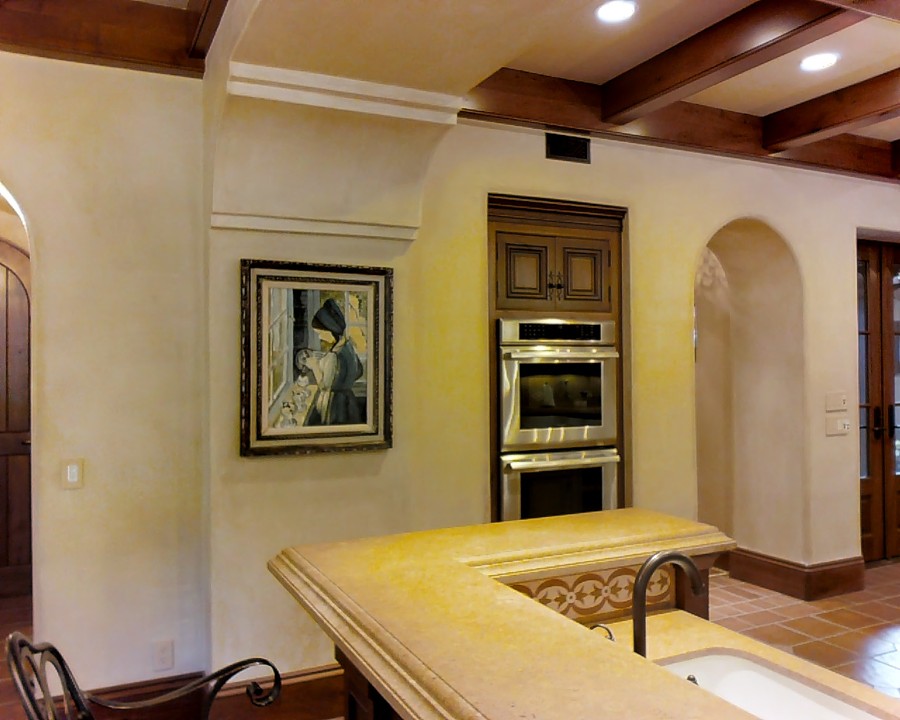}\par
{\footnotesize Image 2}\end{minipage}
\par\medskip
\textbf{Direct}\quad C. Southwest\hfill\textcolor[HTML]{B8493B}{$\times$ Incorrect}\par
\smallskip\textcolor{SBMechanism!30}{\rule{\linewidth}{.35pt}}\par
\textbf{\textcolor{SBMechanism}{SpatialBrain}}\par\smallskip
\begin{minipage}[t]{.46\linewidth}\vspace{0pt}\centering
\includegraphics[width=\linewidth,height=4.2cm,keepaspectratio]{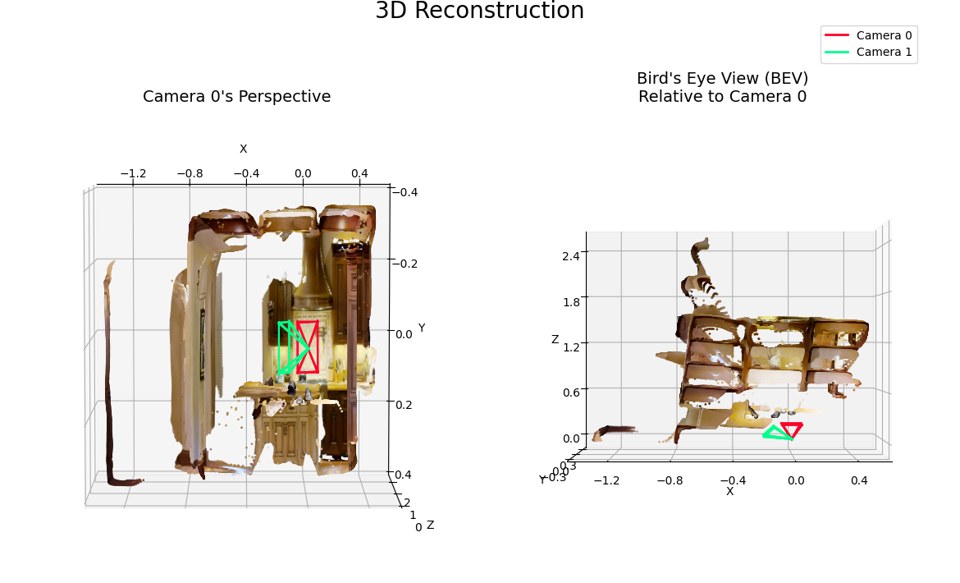}\par
{\footnotesize Reconstruction and camera poses}\par\smallskip
\begin{tcolorbox}[colback=white,colframe=SBMechanism!35,boxrule=.4pt,arc=1mm,left=2mm,right=2mm,top=1mm,bottom=1mm]
{\footnotesize \textbf{Memory available to the planner}\par 4 anchors; 6 relation pairs.\par Atlas includes relation uncertainty.}
\end{tcolorbox}
\end{minipage}\hfill\begin{minipage}[t]{.50\linewidth}\vspace{0pt}
\begin{enumerate}\setlength{\itemsep}{4pt}\setlength{\parskip}{0pt}\setlength{\topsep}{0pt}
\item Project the stove, mural, and two island observations into the reconstructed kitchen. Four anchors and six relation pairs are retained in 3DFusionMemory.
\item Expose the atlas to the planner. The coordinate check combines the two island observations after a consistency check and uses the stove to define east.
\item The mural displacement has east = -0.360 and north = +0.432. Return D: northwest.
\end{enumerate}\end{minipage}\par\medskip
\textbf{Answer}\quad D. Northwest\hfill\textcolor[HTML]{147F77}{$\checkmark$ Correct}\par
\smallskip{\footnotesize\itshape The coordinate check combines the island observations; memory keeps them separate and does not override the answer.}\par
\end{tcolorbox}
\endgroup

\clearpage

\input{sections/appendix_memory_gallery}

\clearpage
\subsection{Spatial Skill Library}
\label{app:cases-skill}
\definecolor{SBCaseAccent}{HTML}{65428A}
These examples show how existing skills provide structured spatial outputs
for answer generation.

\refstepcounter{sbCase}\label{case:mechanism-32}
\begin{sbCompactCase}{Skill Library: Reusable Rotation Estimation\hfill\normalfont\small MMSI 32}
\sbCaseQuestion
The images are taken continuously from a first-person perspective. In which direction are you rotating?\par\smallskip
A. Left\quad B. Right\quad C. Up\quad D. Down

\par\smallskip\noindent
\begin{minipage}[t]{.485\linewidth}\centering
\includegraphics[width=\linewidth,height=2.6cm,keepaspectratio]{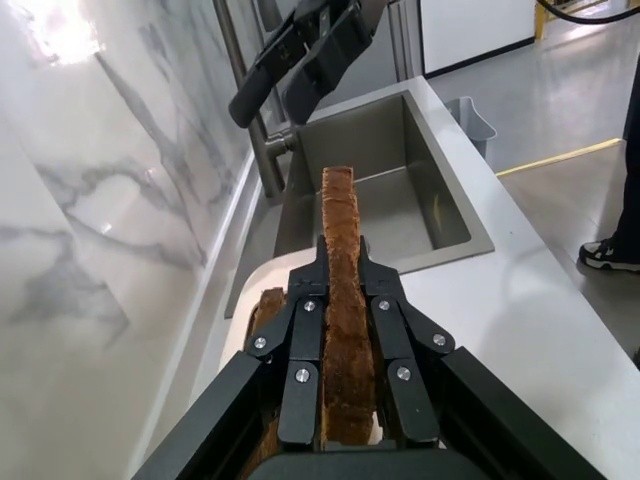}\par
{\footnotesize Image 1}\end{minipage}\hfill
\begin{minipage}[t]{.485\linewidth}\centering
\includegraphics[width=\linewidth,height=2.6cm,keepaspectratio]{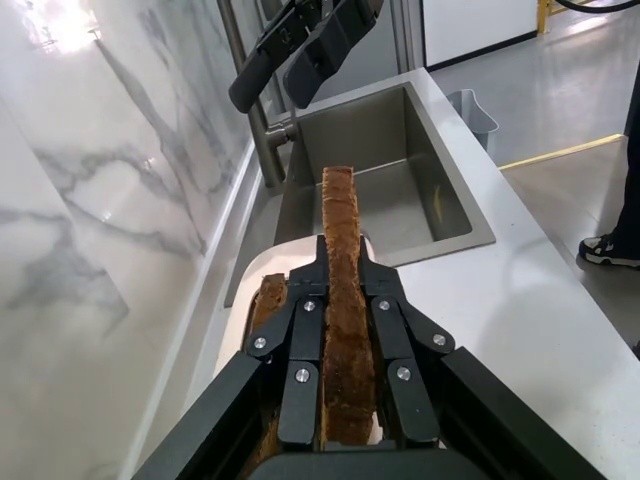}\par
{\footnotesize Image 2}\end{minipage}\par\smallskip
\sbCaseAnswer{Direct}{A. Left}{SBRed}{$\times$ Incorrect}
\sbCaseStep
Invoke \texttt{camera-camera-rotation-template} on the two ordered views.
The skill returns \emph{down}, with primary axis \(x\) and total rotation
1.095 degrees. Final-answer generation uses this returned direction.
\sbCaseAnswer{Answer}{D. Down}{SBGreen}{$\checkmark$ Correct}
{\footnotesize\itshape The existing executable skill is used at its initial state v0.}
\end{sbCompactCase}

\refstepcounter{sbCase}\label{case:mechanism-42}
\begin{sbCompactCase}{Skill Library: Camera-frame Reasoning\hfill\normalfont\small MMSI 42}
\sbCaseQuestion
When you took the photo in Figure 1, where was the iron refrigerator located relative to you?\par\smallskip
A. Front left\quad B. Rear left\quad C. Rear right\quad D. Front right

\par\smallskip\noindent
\begin{minipage}[t]{.485\linewidth}\centering
\includegraphics[width=\linewidth,height=2.6cm,keepaspectratio]{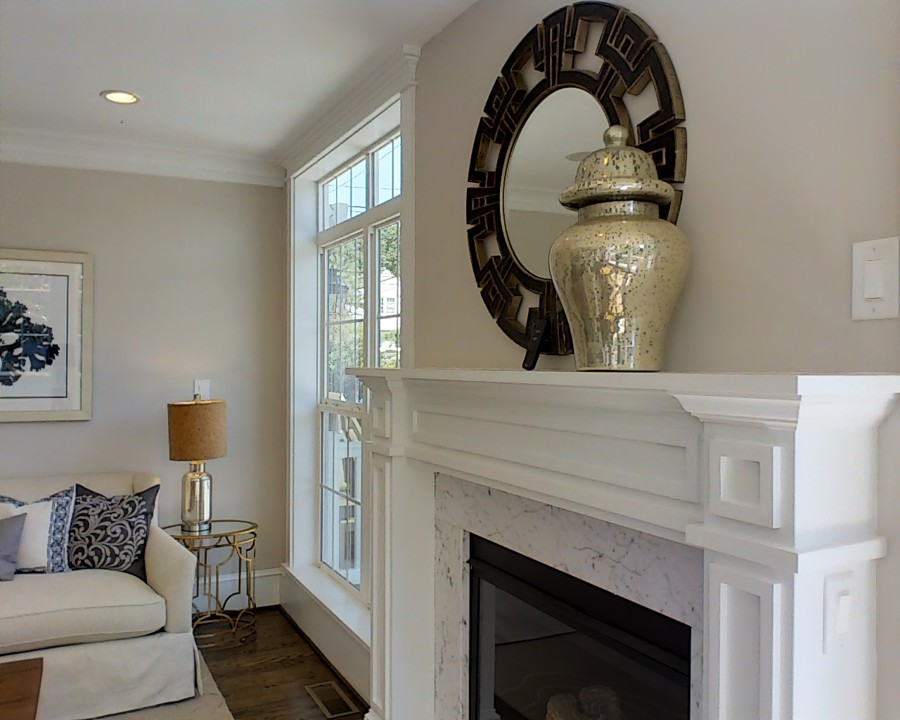}\par
{\footnotesize Image 1}\end{minipage}\hfill
\begin{minipage}[t]{.485\linewidth}\centering
\includegraphics[width=\linewidth,height=2.6cm,keepaspectratio]{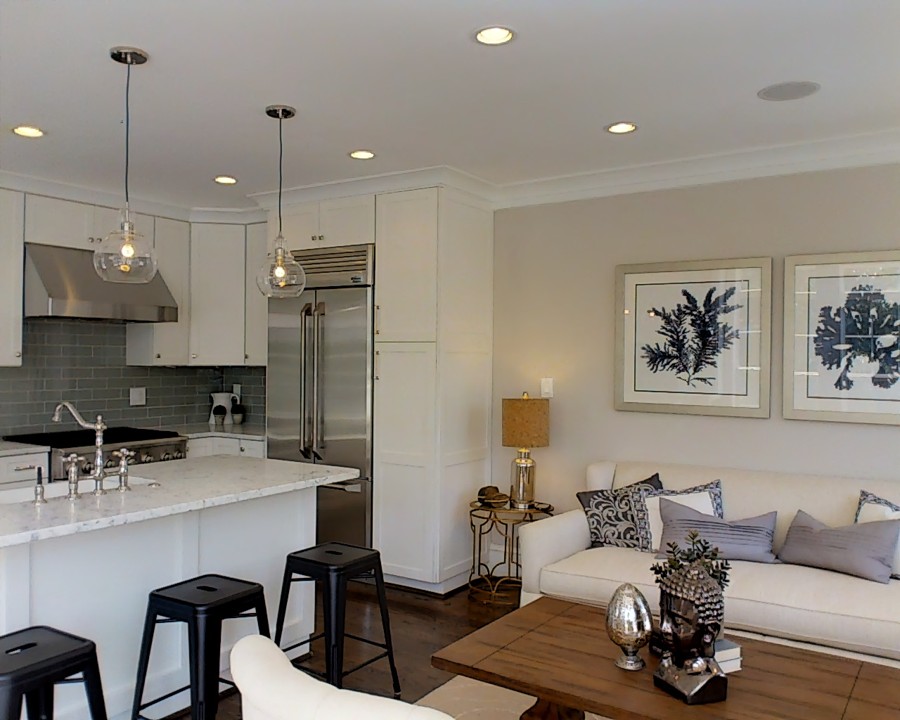}\par
{\footnotesize Image 2}\end{minipage}\par\smallskip
\sbCaseAnswer{Direct}{B. Rear left}{SBRed}{$\times$ Incorrect}
\sbCaseStep
Invoke \texttt{camera-object-direction-template} with camera 0 as the
reference. Its output is \emph{left}, with \(x=-1.377\) and \(z=+0.566\).
These coordinates resolve the diagonal option: left and in front.
\sbCaseAnswer{Answer}{A. Front left}{SBGreen}{$\checkmark$ Correct}
{\footnotesize\itshape The final answer uses the returned coordinates, not just the coarse direction label.}
\end{sbCompactCase}

\clearpage
\subsection{SpatialCognition}
\label{app:cases-routing}
\definecolor{SBCaseAccent}{HTML}{946316}
These cases trace recommendations from earlier task blocks to the current
resource decision. The target is excluded from the supporting experience.

\refstepcounter{sbCase}\label{case:mmsi-748}
\begin{sbCompactCase}{SpatialCognition: Experience-Guided Routing\hfill\normalfont\small MMSI 748}
\sbCaseQuestion
Which is longer: the long side of the statue's base in Figure 1 or the width of the bench in Figure 2?\par\smallskip
{\footnotesize
A. They are the same length\par
B. The long side of the statue's base in Figure 1 is longer\par
C. Sometimes the long side of the statue's base in Figure 1 is longer, sometimes the width of the bench in Figure 2 is longer\par
D. The width of the bench in Figure 2 is longer}

\par\smallskip\noindent
\begin{minipage}[t]{.485\linewidth}\centering
\includegraphics[width=\linewidth,height=2.4cm,keepaspectratio]{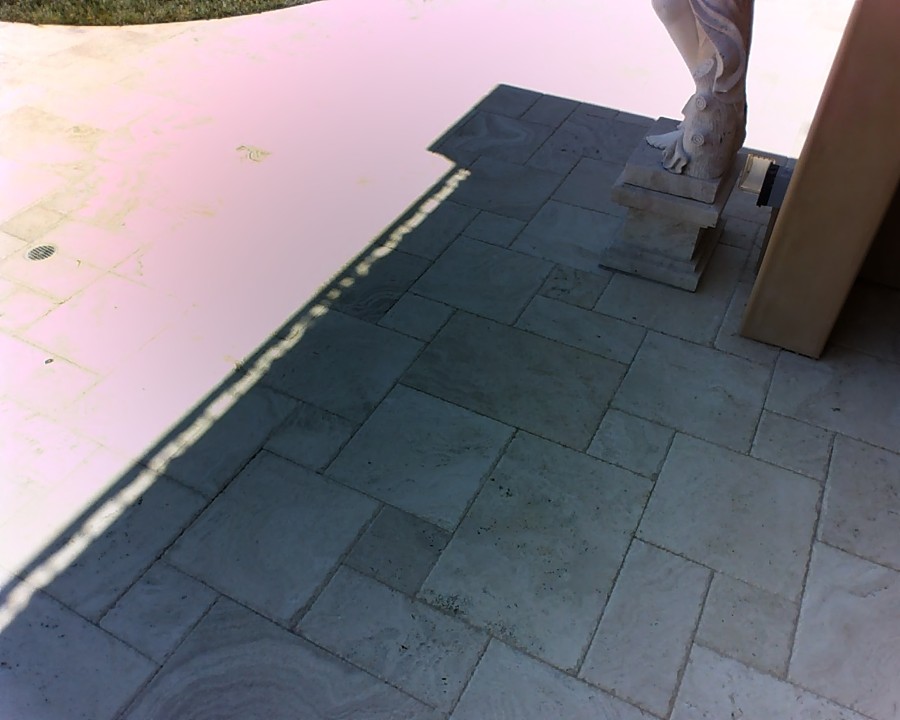}\par
{\footnotesize Image 1}\end{minipage}\hfill
\begin{minipage}[t]{.485\linewidth}\centering
\includegraphics[width=\linewidth,height=2.4cm,keepaspectratio]{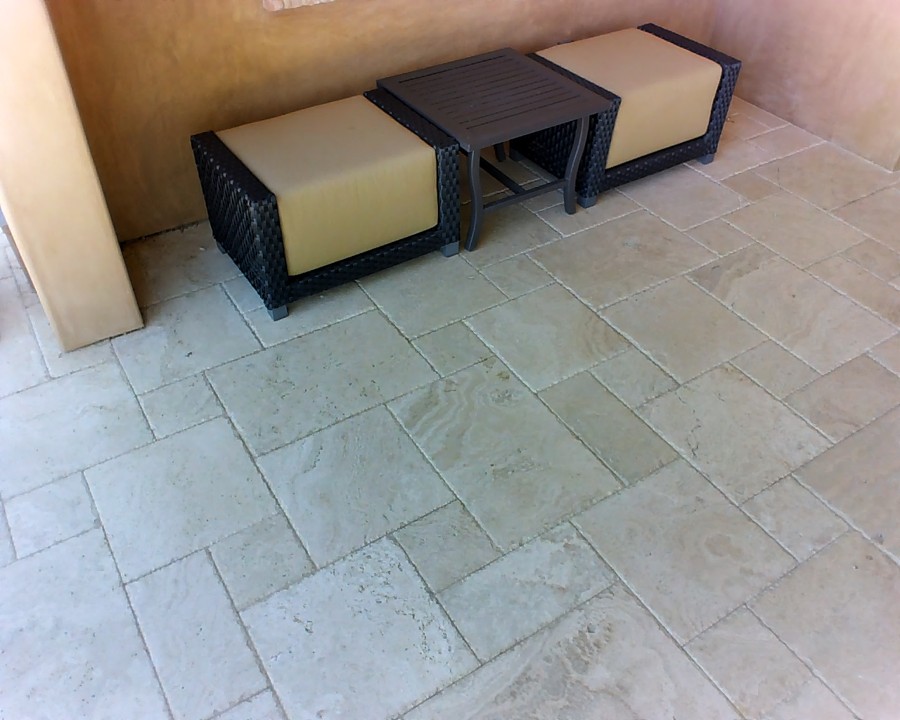}\par
{\footnotesize Image 2}\end{minipage}\par\smallskip
\sbCaseAnswer{Direct}{D. The width of the bench in Figure 2 is longer}{SBGreen}{$\checkmark$ Correct}
\sbCaseStep
\textbf{Earlier experience:} 27 same-family events.
\textbf{Policy change:} v10 recommends agent execution; v11 recommends direct
reasoning. The current task follows v11 without exploration, with memory
off and no skill execution. The 2D-measurement guard is shared by both arms
and remains unchanged.
\sbCaseAnswer{Answer}{D. The width of the bench in Figure 2 is longer}{SBGreen}{$\checkmark$ Correct}
{\footnotesize\textbf{Fixed-policy control \(\rightarrow\) SpatialBrain (both correct):}
37,884 \(\rightarrow\) 3,632 tokens (\(-90.4\%\));
2 \(\rightarrow\) 0 non-final top-level calls;
38.0 \(\rightarrow\) 9.4\,s observed.\par
Standalone Direct also answers correctly: 3,697 versus 3,632 tokens
(\(-1.8\%\)); both use zero tools.}
\end{sbCompactCase}

\refstepcounter{sbCase}\label{case:mmsi-348}
\begin{sbCompactCase}{SpatialCognition: Adaptive Skill Selection\hfill\normalfont\small MMSI 348}
\sbCaseQuestion
What kind of angle is approximately formed by the projections on the ground of the forward directions of the camera when taking the first and second pictures?\par\smallskip
A. Obtuse angle\quad B. 0 degrees\quad C. Right angle\quad D. Acute angle

\par\smallskip\noindent
\begin{minipage}[t]{.485\linewidth}\centering
\includegraphics[width=\linewidth,height=2.2cm,keepaspectratio]{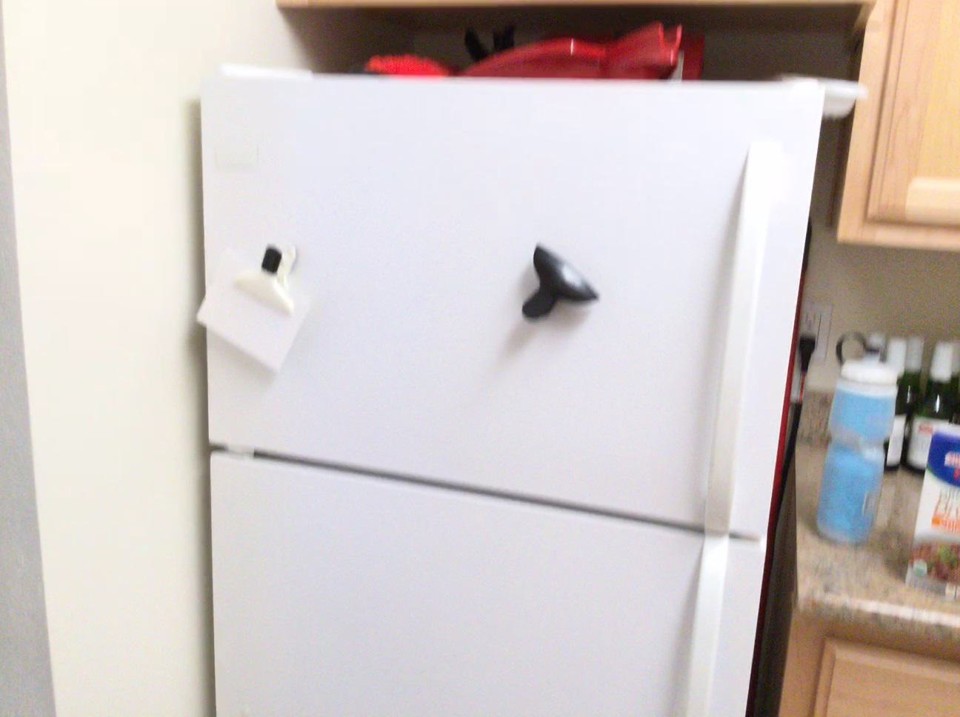}\par
{\footnotesize Image 1}\end{minipage}\hfill
\begin{minipage}[t]{.485\linewidth}\centering
\includegraphics[width=\linewidth,height=2.2cm,keepaspectratio]{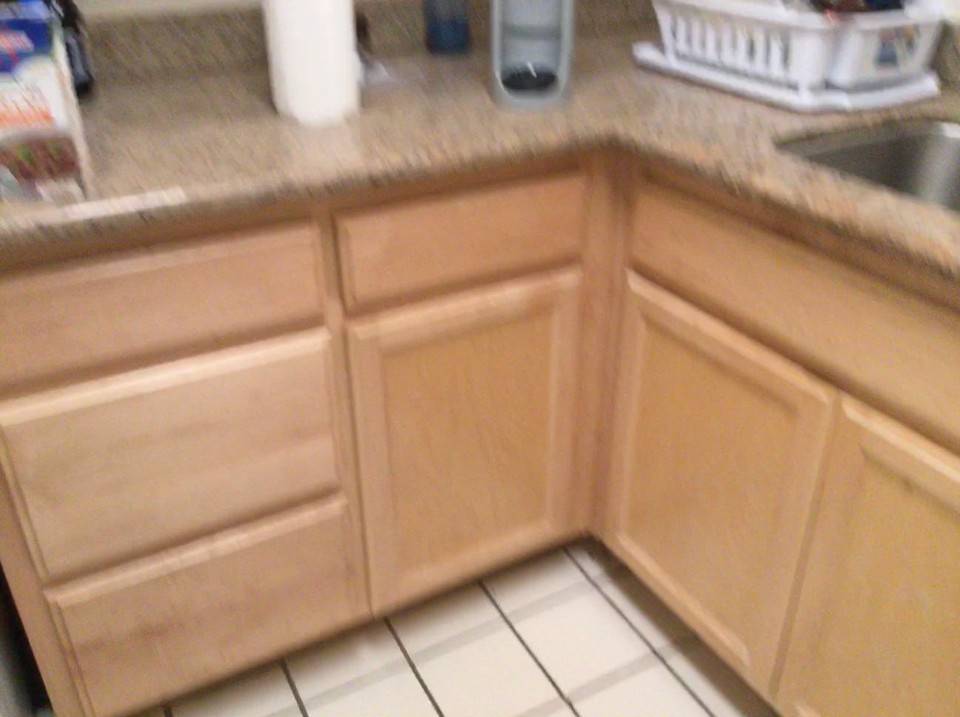}\par
{\footnotesize Image 2}\end{minipage}\par\smallskip
\sbCaseAnswer{Direct}{D. Acute angle}{SBGreen}{$\checkmark$ Correct}
\sbCaseStep
Earlier experience changes the recommendation for the existing
\texttt{camera-camera-rotation-template} from hint-only (v1) to executable
(v4). With 18 earlier same-family events, the current v5 policy executes
the skill without exploration. Its returned angle is 47.6 degrees.
\sbCaseAnswer{Answer}{D. Acute angle}{SBGreen}{$\checkmark$ Correct}
{\footnotesize\itshape The fixed-policy control also calls this skill.
The observed change is in the selection policy, not the skill code or this item's correctness.}
\end{sbCompactCase}

\clearpage
\subsection{Additional Execution Traces}
\label{app:cases-tools}
\definecolor{SBCaseAccent}{HTML}{486A83}
The following paths illustrate coordinate conventions, temporal
correspondence, and execution costs. They complement the component examples
without attributing each tool decision to accumulated experience.

\refstepcounter{sbCase}\label{case:mmsi-18}
\begin{sbCompactCase}{Tool Trajectory: Coordinate-convention Alignment\hfill\normalfont\small MMSI 18}
\sbCaseQuestion
The camera coordinate system is defined as +Y up, -Z forward, and is a right-handed system. How is the viewpoint in the second image obtained from the viewpoint in the first image by moving in its corresponding camera coordinate system?\par\smallskip
{\footnotesize
A. Move in the positive Z direction and positive Y direction\par
B. Move in the negative Z direction and positive Y direction\par
C. Move in the negative X direction and negative Y direction\par
D. Move in the positive X direction and negative Y direction}

\par\smallskip\noindent
\begin{minipage}[t]{.485\linewidth}\centering
\includegraphics[width=\linewidth,height=2.4cm,keepaspectratio]{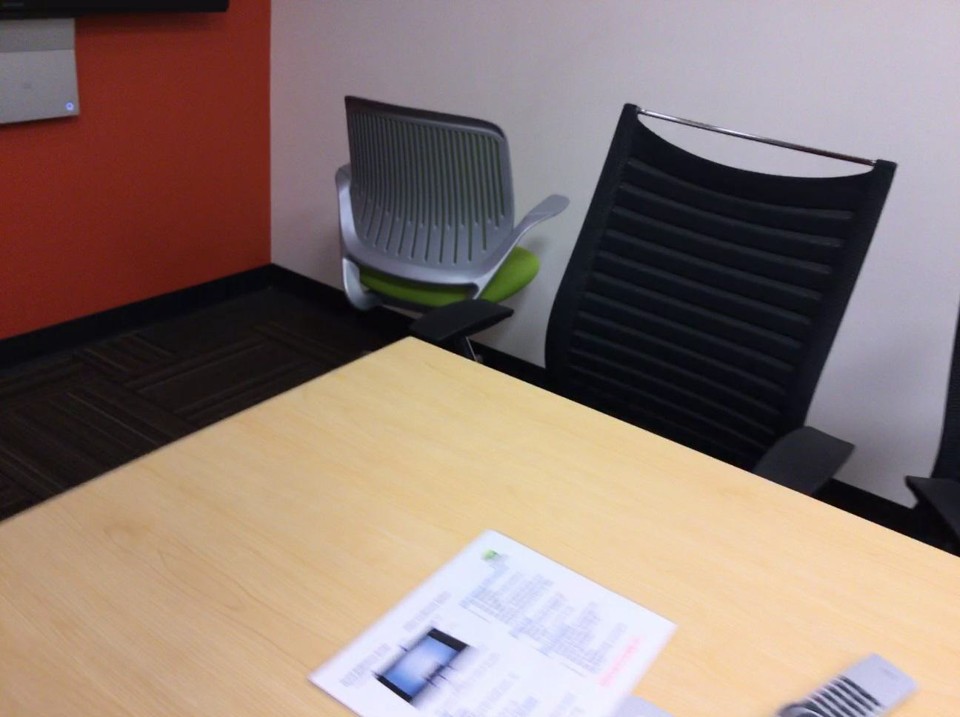}\par
{\footnotesize Image 1}\end{minipage}\hfill
\begin{minipage}[t]{.485\linewidth}\centering
\includegraphics[width=\linewidth,height=2.4cm,keepaspectratio]{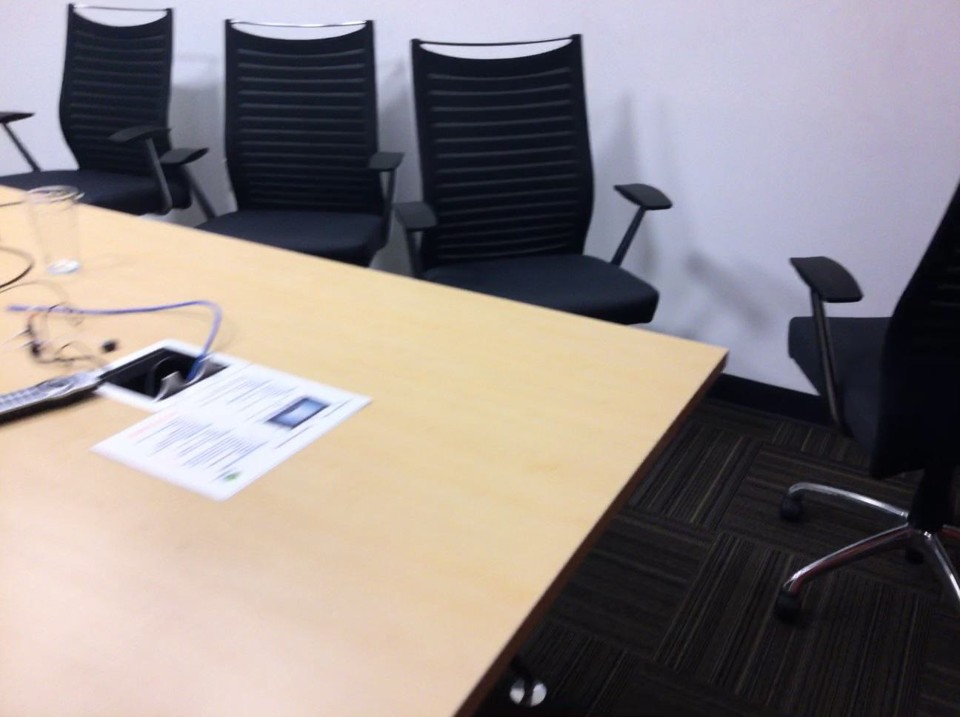}\par
{\footnotesize Image 2}\end{minipage}\par\smallskip
\sbCaseAnswer{Direct}{D. Move in the positive X direction and negative Y direction}{SBRed}{$\times$ Incorrect}
\sbCaseStep
Reconstruct the views and recover their camera centers. Express the second
center in the first camera frame under the specified convention.
The displacement \((-0.083,+0.509,-1.060)\) gives negative Z and positive Y.
\sbCaseAnswer{Answer}{B. Move in the negative Z direction and positive Y direction}{SBGreen}{$\checkmark$ Correct}
\end{sbCompactCase}

\refstepcounter{sbCase}\label{case:mmsi-82}
\begin{sbCompactCase}{Tool Trajectory: Temporal Correspondence\hfill\normalfont\small MMSI 82}
\sbCaseQuestion
The pictures were taken in sequence. In which direction are the two grippers moving?\par\smallskip
{\footnotesize
A. The left gripper did not move, the right gripper moved to the left\par
B. The left gripper moved to the right, the right gripper moved to the right\par
C. The left gripper moved to the right, the right gripper did not move\par
D. The left gripper moved to the left, the right gripper did not move.}

\par\smallskip\noindent
\begin{minipage}[t]{.485\linewidth}\centering
\includegraphics[width=\linewidth,height=2.4cm,keepaspectratio]{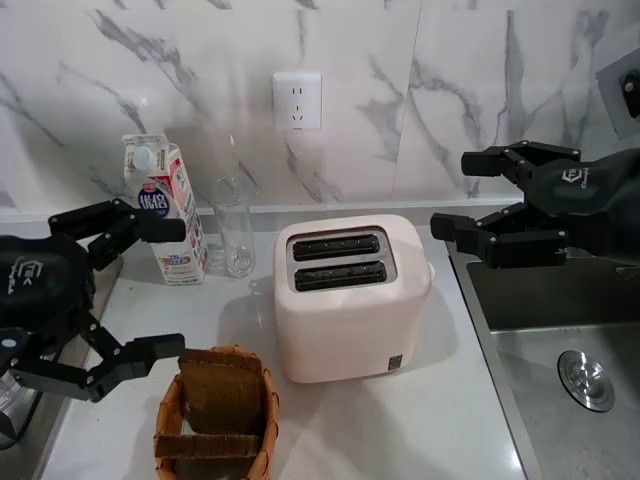}\par
{\footnotesize Image 1}\end{minipage}\hfill
\begin{minipage}[t]{.485\linewidth}\centering
\includegraphics[width=\linewidth,height=2.4cm,keepaspectratio]{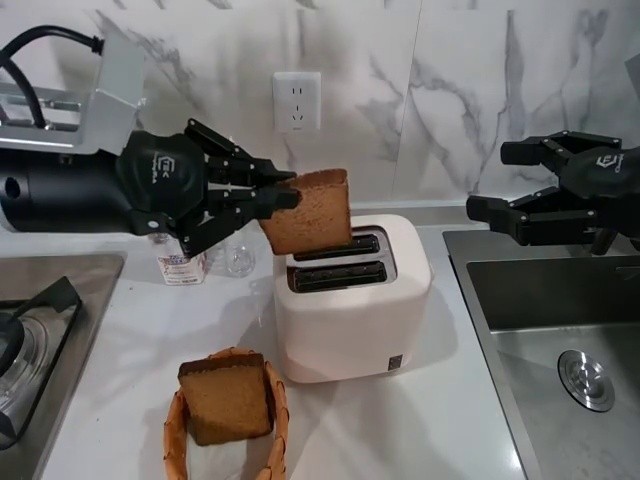}\par
{\footnotesize Image 2}\end{minipage}\par\smallskip
\sbCaseAnswer{Direct}{C. The left gripper moved to the right, the right gripper did not move}{SBRed}{$\times$ Incorrect}
\sbCaseStep
Detect both grippers and select corresponding boxes across frames.
Their horizontal centers shift by +57.3 and +17.7 pixels, exceeding the
respective stationary thresholds of 9.8 and 7.7 pixels.
\sbCaseAnswer{Answer}{B. The left gripper moved to the right, the right gripper moved to the right}{SBGreen}{$\checkmark$ Correct}
{\footnotesize\itshape Multiple detections are present; the calculation uses the selected correspondences.}
\end{sbCompactCase}

\clearpage
\refstepcounter{sbCase}\label{case:mmsi-289}
\begin{sbCompactCase}{Tool Trajectory: Reduced Tool Use\hfill\normalfont\small MMSI 289}
\sbCaseQuestion
The images are taken continuously from a first-person perspective. In which direction are you moving?\par\smallskip
A. Front left\quad B. Front right\quad C. Right\quad D. Forward

\par\smallskip\noindent
\begin{minipage}[t]{.485\linewidth}\centering
\includegraphics[width=\linewidth,height=2.6cm,keepaspectratio]{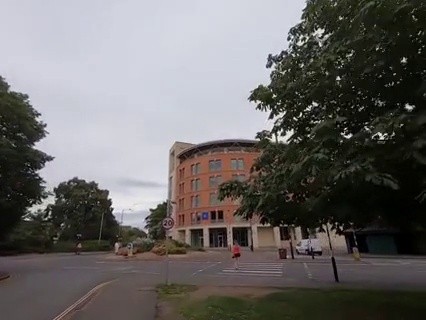}\par
{\footnotesize Image 1}\end{minipage}\hfill
\begin{minipage}[t]{.485\linewidth}\centering
\includegraphics[width=\linewidth,height=2.6cm,keepaspectratio]{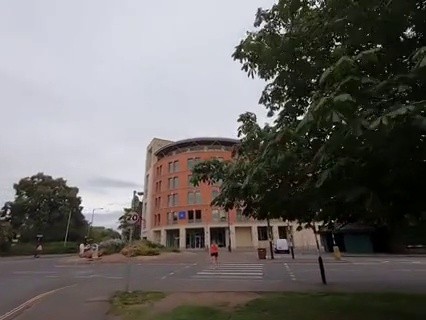}\par
{\footnotesize Image 2}\end{minipage}\par\smallskip
\sbCaseAnswer{Direct}{C. Right}{SBRed}{$\times$ Incorrect}
\sbCaseStep
Request landmark detections for the brick building and roadside sign in
both frames. Proceed from the two detection calls without a reconstruction
or Python call. The fixed-policy run reaches the same answer after an
additional reconstruction and two Python calls.
\sbCaseAnswer{Answer}{B. Front right}{SBGreen}{$\checkmark$ Correct}
{\footnotesize\textbf{Fixed-policy control \(\rightarrow\) SpatialBrain (both correct):}
88,760 \(\rightarrow\) 29,415 tokens (\(-66.9\%\));
5 \(\rightarrow\) 2 non-final top-level calls;
318.6 \(\rightarrow\) 29.7\,s observed.}
\end{sbCompactCase}

\refstepcounter{sbCase}\label{case:mmsi-880}
\begin{sbCompactCase}{Tool Trajectory: Skill Fallback and Recovery\hfill\normalfont\small MMSI 880}
\sbCaseQuestion
The picture is a continuous first-person perspective shot. In which direction is the spoon moving?\par\smallskip
A. Out of the bowl\quad B. Toward the person\quad C. Into the bowl\quad D. Toward the pot

\par\smallskip\noindent
\begin{minipage}[t]{.485\linewidth}\centering
\includegraphics[width=\linewidth,height=2.6cm,keepaspectratio]{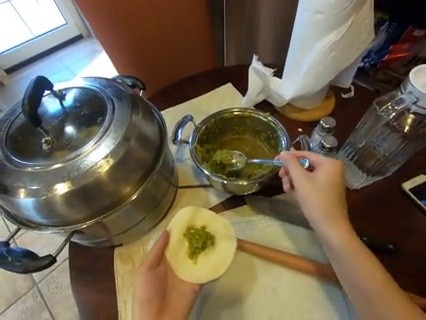}\par
{\footnotesize Image 1}\end{minipage}\hfill
\begin{minipage}[t]{.485\linewidth}\centering
\includegraphics[width=\linewidth,height=2.6cm,keepaspectratio]{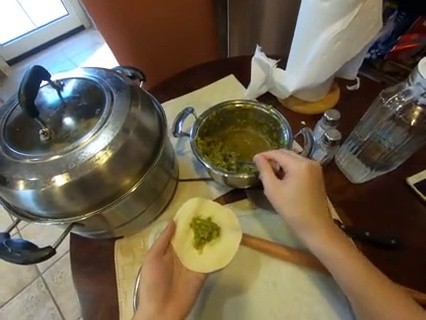}\par
{\footnotesize Image 2}\end{minipage}\par\smallskip
\sbCaseAnswer{Direct}{C. Into the bowl}{SBGreen}{$\checkmark$ Correct}
\sbCaseStep
The object-motion skill returns control to the planner because slot
information is insufficient; no nested tool executes. The planner then
makes one spoon-detection call per frame. The fixed-policy run encounters
the same fallback but makes eight detector calls.
\sbCaseAnswer{Answer}{C. Into the bowl}{SBGreen}{$\checkmark$ Correct}
{\footnotesize\textbf{Fixed-policy control \(\rightarrow\) SpatialBrain (both correct):}
90,898 \(\rightarrow\) 52,417 tokens (\(-42.3\%\));
9 \(\rightarrow\) 3 non-final top-level calls;
110.5 \(\rightarrow\) 69.4\,s observed.\par
Both runs encounter ambiguous detections. Counts include the fallback;
this is recovery after an unsuccessful skill attempt.}
\end{sbCompactCase}

\clearpage
\subsubsection*{Reasoning and Cost Boundaries}
Correct answers can coexist with an unsupported geometric shortcut or a
more expensive execution path. These cases distinguish answer correctness
from the quality and cost of the supporting procedure.

\refstepcounter{sbCase}\label{case:mmsi-715}
\begin{sbCompactCase}{Boundary Case: Cross-view Height Comparison\hfill\normalfont\small MMSI 715}
\sbCaseQuestion
Which is higher, the fabric on the left or the sink on the right?\par\smallskip
{\footnotesize A. The same height\quad B. The sink\quad C. The fabric\par
D. Sometimes the fabric is higher, sometimes the sink is higher}

\par\smallskip\noindent
\begin{minipage}[t]{.485\linewidth}\centering
\includegraphics[width=\linewidth,height=2.5cm,keepaspectratio]{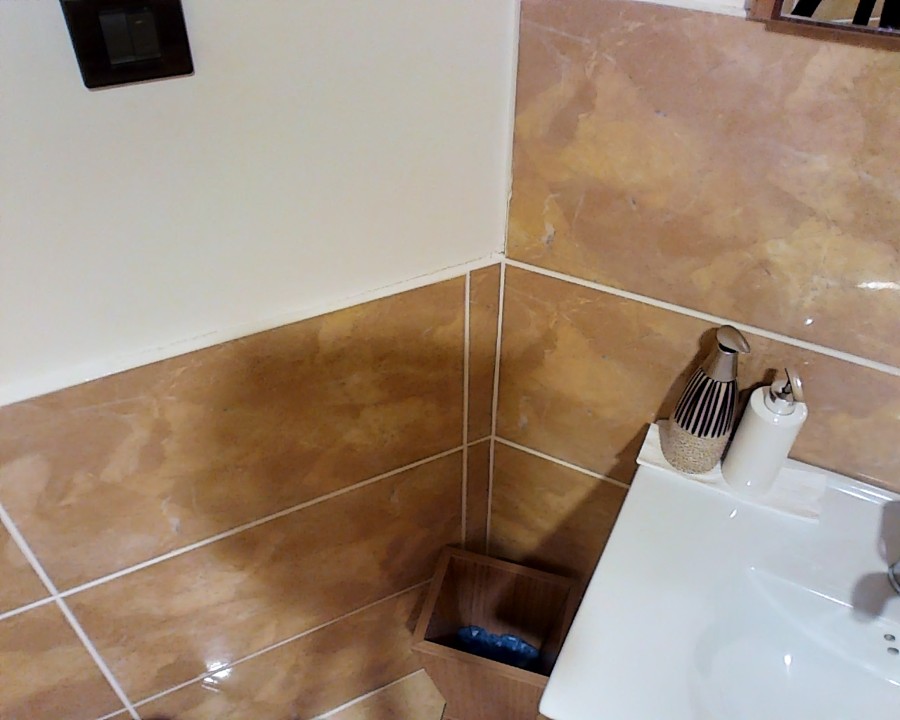}\par
{\footnotesize Image 1}\end{minipage}\hfill
\begin{minipage}[t]{.485\linewidth}\centering
\includegraphics[width=\linewidth,height=2.5cm,keepaspectratio]{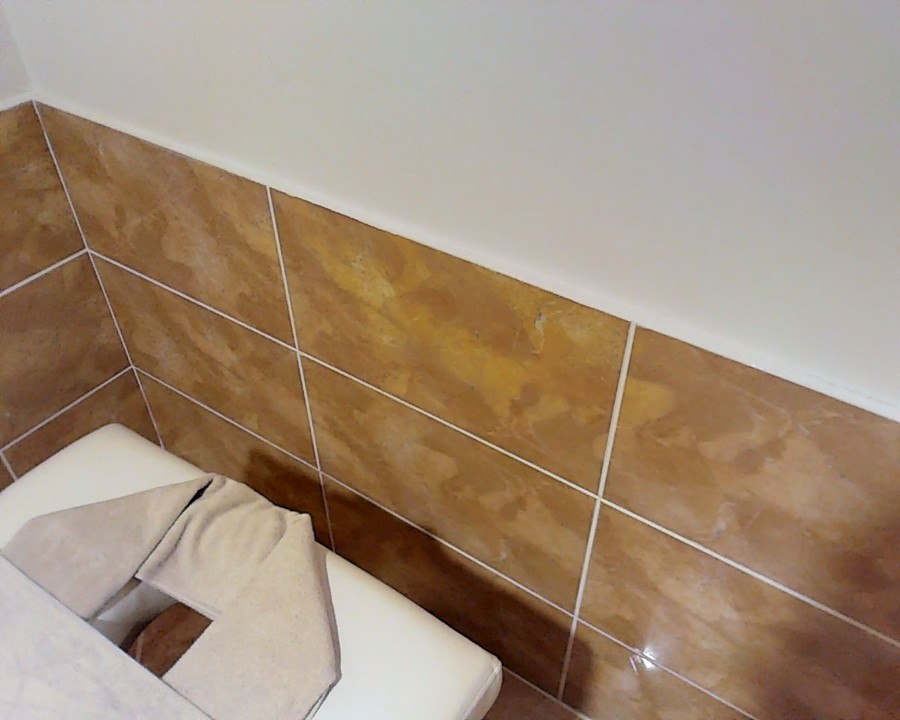}\par
{\footnotesize Image 2}\end{minipage}\par\smallskip
\sbCaseAnswer{Direct}{B. The sink}{SBGreen}{$\checkmark$ Correct}
\sbCaseAnswer{SpatialBrain}{B. The sink}{SBGreen}{$\checkmark$ Correct}
\smallskip
The procedure detects the sink in image 1 and the fabric in image 2, then
compares normalized box-center heights, 0.792 and 0.825.
Its description invokes shared wall tiles as context, but the computation
does not estimate a cross-view alignment. Normalization alone does not
establish a common height reference across these views; the correct answer
therefore does not validate this geometric shortcut.
\par\smallskip{\footnotesize Both runs use the same memory-off/hint route.
Fixed-policy control \(\rightarrow\) SpatialBrain:
64,712 \(\rightarrow\) 58,586 tokens (\(-9.5\%\)),
3 \(\rightarrow\) 3 non-final top-level calls, and
196.1 \(\rightarrow\) 102.6\,s observed.
Standalone Direct is also correct with 3,534 tokens.}
\end{sbCompactCase}

\refstepcounter{sbCase}\label{case:mmsi-416}
\begin{sbCompactCase}{Boundary Case: Compute Trade-offs\hfill\normalfont\small MMSI 416}
\sbCaseQuestion
This is a set of photos I took at home. In the last photo, in which direction is the exit area relative to me?\par\smallskip
A. front left\quad B. back left\quad C. front right\quad D. back right

\par\smallskip\noindent
\begin{minipage}[t]{.485\linewidth}\centering
\includegraphics[width=\linewidth,height=2.5cm,keepaspectratio]{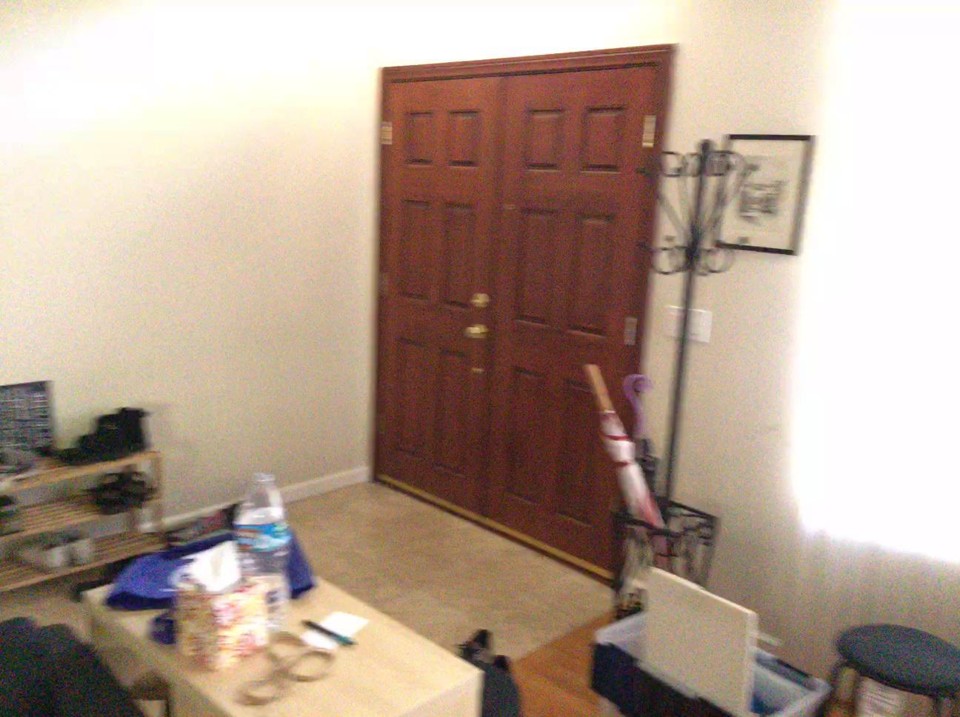}\par
{\footnotesize Image 1}\end{minipage}\hfill
\begin{minipage}[t]{.485\linewidth}\centering
\includegraphics[width=\linewidth,height=2.5cm,keepaspectratio]{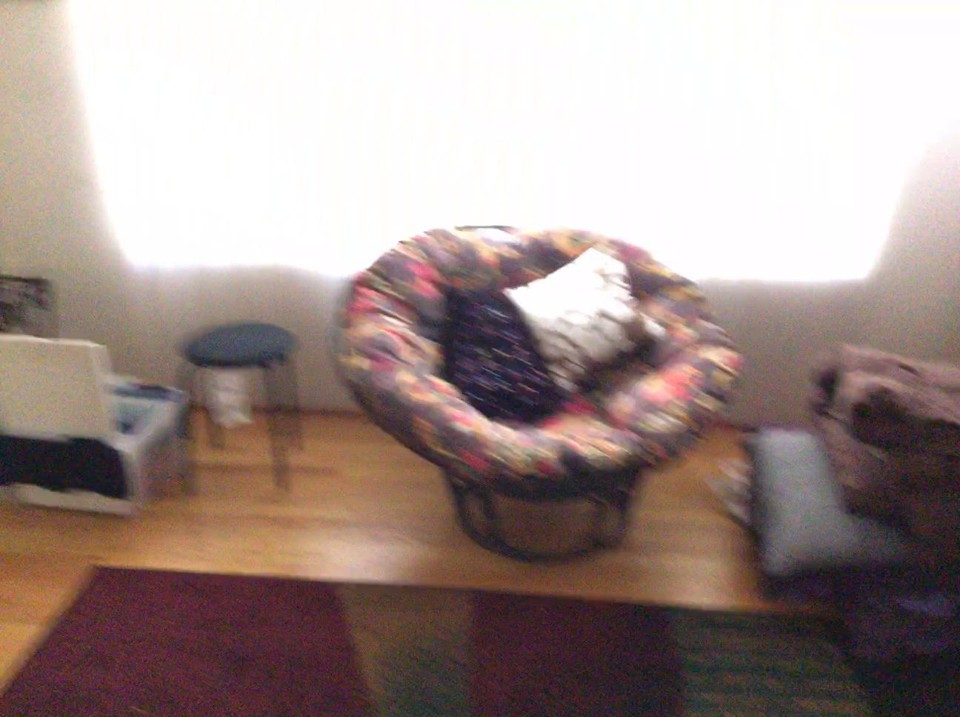}\par
{\footnotesize Image 2}\end{minipage}\par\smallskip
\sbCaseAnswer{Direct}{A. front left}{SBGreen}{$\checkmark$ Correct}
\sbCaseAnswer{SpatialBrain}{A. front left}{SBGreen}{$\checkmark$ Correct}
\smallskip
Detect the exit, reconstruct both views, and project the exit into 3D.
Its coordinates in the last camera frame are \(x=-0.852,\ z=+0.987\),
giving front left. This is an exploratory route with memory disabled.
\par\smallskip{\footnotesize Fixed-policy control \(\rightarrow\) SpatialBrain:
123,214 \(\rightarrow\) 111,609 tokens (\(-9.4\%\)),
4 \(\rightarrow\) 5 non-final top-level calls, and
258.9 \(\rightarrow\) 262.3\,s observed.
Fewer tokens here do not imply fewer calls or shorter time.}
\end{sbCompactCase}
\endgroup

%% file: sections/appendix_memory_gallery.tex
\clearpage
\subsubsection*{Multi-view Support in 3DFusionMemory}
These two examples connect image detections to 3D anchors, camera poses, and
supporting views. We selected them for visible geometry, without using answer
correctness. The overhead inset shows the same state from a second viewpoint,
not a later fusion step. Colored matrix cells indicate recorded view support,
not independently verified object matches. All coordinates are unchanged;
distance scales are not calibrated in meters. Dashed lines mark uncertain
relations, and each diagram shows a subset of the stored relations.
\begingroup
\newcounter{sbMemoryCase}
\renewcommand{\thesbMemoryCase}{\ifnum\value{sbMemoryCase}<10 0\fi\arabic{sbMemoryCase}}
\definecolor{SBMemoryTeal}{HTML}{207C75}
\definecolor{SBMemoryPale}{HTML}{F3F8F7}
\begin{tcolorbox}[enhanced,breakable,
colback=SBMemoryPale,colframe=SBMemoryTeal,colbacktitle=SBMemoryTeal,coltitle=white,
boxrule=.7pt,arc=2mm,left=3mm,right=3mm,top=2mm,bottom=3mm,fonttitle=\bfseries,
title={3DFusionMemory: Spatial Representation\hfill\normalfont\small Memory 01}]
\refstepcounter{sbMemoryCase}
\label{case:memory-mmsi-297}
{\footnotesize Representation Example $\cdot$ MMSI 297}\par\smallskip
\textbf{\textcolor{SBMemoryTeal}{Objects supported across views}}\par\smallskip
{\small 6 camera views $\cdot$ 5 3D anchors $\cdot$ 10 stored relation pairs}\par\smallskip
\noindent\makebox[\linewidth][c]{\includegraphics[width=.94\linewidth]{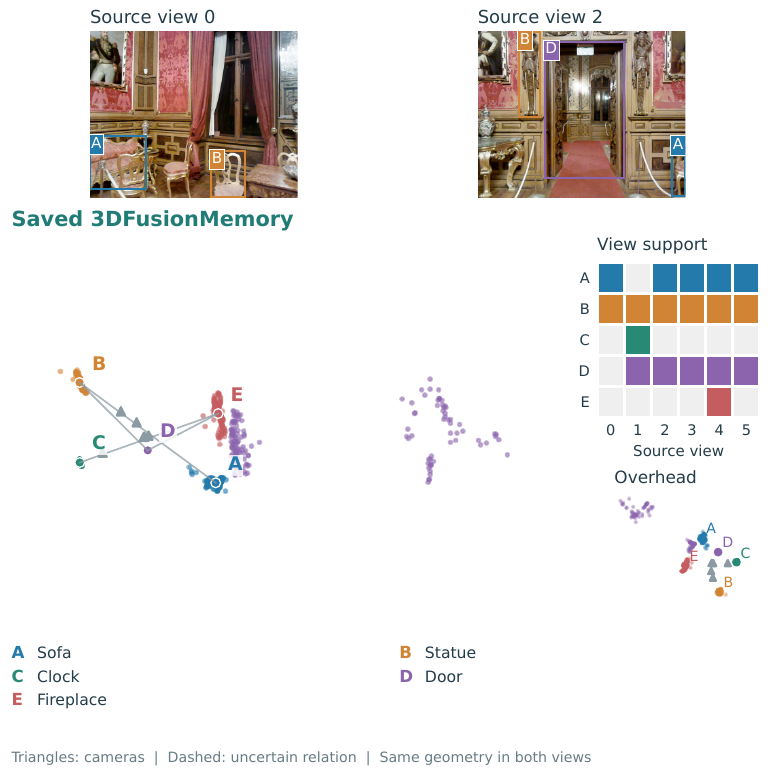}}
\par\smallskip{\small The sofa, statue, and door retain support from multiple views. The clock and fireplace have single-view support. Shared letter IDs connect detections, saved 3D anchors, and the support matrix.}
\par\smallskip{\footnotesize 3/5 anchors have support in at least two views.}
\end{tcolorbox}
\clearpage
\begin{tcolorbox}[enhanced,breakable,
colback=SBMemoryPale,colframe=SBMemoryTeal,colbacktitle=SBMemoryTeal,coltitle=white,
boxrule=.7pt,arc=2mm,left=3mm,right=3mm,top=2mm,bottom=3mm,fonttitle=\bfseries,
title={3DFusionMemory: Spatial Representation\hfill\normalfont\small Memory 02}]
\refstepcounter{sbMemoryCase}
\label{case:memory-mmsi-861}
{\footnotesize Representation Example $\cdot$ MMSI 861}\par\smallskip
\textbf{\textcolor{SBMemoryTeal}{Mixed support in an indoor scene}}\par\smallskip
{\small 6 camera views $\cdot$ 6 3D anchors $\cdot$ 15 stored relation pairs}\par\smallskip
\noindent\includegraphics[width=.90\linewidth]{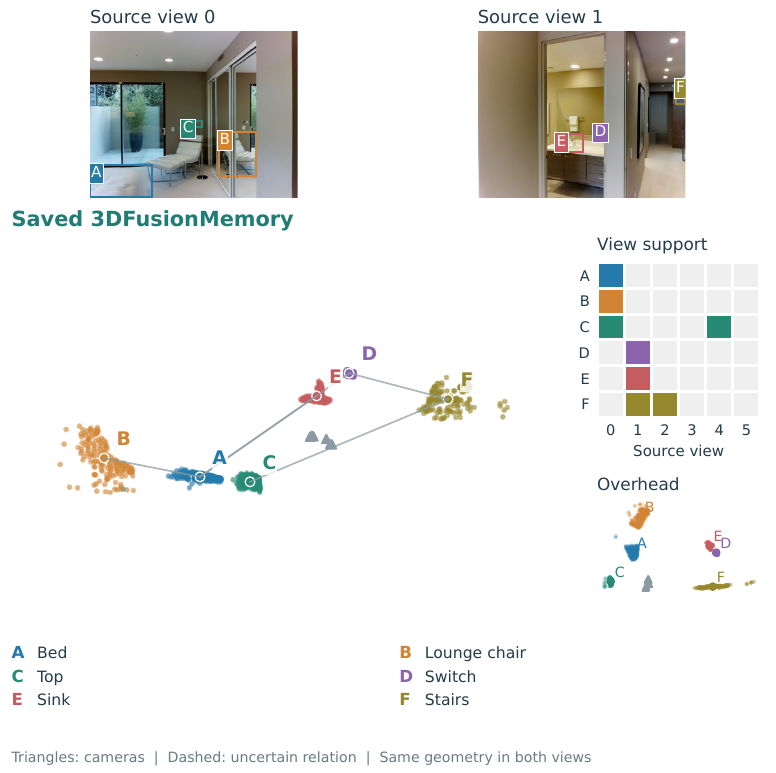}
\par\smallskip{\small The top and stairs retain two-view support; the other four anchors are supported by one view each. The additional overhead view separates the stored layout from the display camera perspective.}
\par\smallskip{\footnotesize 2/6 anchors have support in at least two views.}
\end{tcolorbox}

{\small
\textbf{Single-view support in other saved scenes.}
Four additional scene states distinguish shared-frame organization from
repeated observations of the same object. MMSI 700 and 711 have 8 and 7
anchors, respectively; MindCube around/033d379d and around/0d8a8816 each
have 5. None of these anchors has support in two or more views. In
around/033d379d, all five anchors come from view 1 despite three retained
camera poses. These states illustrate stored spatial layouts rather than
repeated-observation fusion.}
\endgroup

%% file: sections/appendix_experience_updates.tex
\clearpage
\section{Memory Fusion and Experience Updates}
\label{app:experience-update-details}
\subsection{Confidence-weighted Scene Fusion}
\label{app:scene-fusion-details}
Reconstruction provides a common scene frame. For a matched object $o$,
$\boldsymbol\mu_o\in\mathbb R^3$ is its current center and $W_o$ its
accumulated confidence weight. A projection from view $v$ supplies center
$\widetilde{\boldsymbol\mu}_{o,v}$ and weight $w_{o,v}$:
\begin{equation}
\boldsymbol\mu_o^{+}
=\frac{W_o\boldsymbol\mu_o+w_{o,v}\widetilde{\boldsymbol\mu}_{o,v}}
       {W_o+w_{o,v}},\qquad
W_o^{+}=W_o+w_{o,v}.
\label{eq:center-fusion}
\end{equation}
The superscript $+$ denotes the updated state. The view-specific center is
the confidence-weighted centroid of projected points, or their arithmetic
mean if total confidence is zero; $w_{o,v}$ is their mean confidence.
The first valid projection initializes the estimate.
\Eqref{eq:center-fusion} requires $W_o+w_{o,v}>0$. These weights express
relative support, not calibrated probabilities.

\subsection{Conservative Strategy Admission}
\label{app:strategy-admission-details}
In the executable configuration, consider one route or memory factor with
default action $a_0$ and alternative $a$. Statistics are conditioned on the
task signature and use only earlier blocks. Eligibility requires
\begin{equation}
\begin{aligned}
\min\{n_a,n_{a_0}\}&\ge n_{\min},\qquad
n^{\mathrm{cost}}_u=n_u\ \ (u\in\{a,a_0\}),\\
R^{\mathrm{upper}}_a&\le\epsilon,\qquad
C^{\mathrm{upper}}_a\le(1-\delta)C^{\mathrm{lower}}_{a_0}.
\end{aligned}
\label{eq:policy-admission}
\end{equation}
For action $u$, $n_u$ counts observations and $n_u^{\mathrm{cost}}$ complete
cost records; $n_{\min}$ is the minimum support. Incomplete costs cannot
establish a saving. $R_a^{\mathrm{upper}}$ is a Wilson upper estimate of the
process-violation rate, including unverified actions; $\epsilon$ is its
tolerance. The cost bounds use the sample mean plus or minus 1.96 standard
errors, with the lower bound floored at zero. $\delta\in[0,1)$ is the required
fractional saving. Thus \eqref{eq:policy-admission} compares a candidate's
upper cost with the default's lower cost, not just two point estimates.

The controller checks the specific context before a broader task family;
supported context-specific risk can veto a broader recommendation. At each
level it chooses the eligible alternative with the lowest upper cost,
retaining the default if none qualifies. These are process-feedback gates,
not correctness rewards or guarantees of answer accuracy. They describe
the executable configuration, not every historical run in the paper.

\subsection{Learned and Fixed State}
\label{app:learned-fixed-state}
The initial library, task-signature schema, tool interfaces, validation
checks, and update rules are specified before evaluation. Experience changes
support statistics, associations between task conditions and strategies,
and procedural availability. The executable configuration can also compile
and validate new procedures. Some failure-check text comes from predefined
rules: observing a failure adds its applicability and support, rather than
inventing an unrestricted instruction. A selected skill, an exposed hint,
an executed procedure, a verified return, and a correct answer are distinct
events. Process observations cannot establish answer correctness without
the separate evaluation labels, which are not used by the update loop.